\documentclass[preprint]{elsarticle}

\usepackage{amsmath,amssymb,amsthm}
\usepackage[utf8]{inputenc}
\usepackage[english]{babel}
\usepackage{enumitem}
\usepackage{graphicx}
\usepackage{hyperref}
\usepackage{cleveref}
\usepackage{booktabs}
\usepackage{caption}
\usepackage[table]{xcolor}
\usepackage[title,titletoc]{appendix}

\DeclareMathOperator{\tnorm}{\mathcal{T}}
\DeclareMathOperator{\tconorm}{\mathcal{S}}
\DeclareMathOperator{\negator}{\mathcal{N}}
\DeclareMathOperator{\implicator}{\mathcal{I}}

\DeclareMathOperator{\OWA}{OWA}

\DeclareMathOperator{\att}{\mathcal{A}}
\DeclareMathOperator{\train}{train}
\DeclareMathOperator{\test}{test}

\DeclareMathOperator{\CSMBR}{CSMBR}

\newtheorem{example}{Example}

\newcommand{\code}{\texttt}

\begin{document}

\begin{frontmatter}

\title{Evaluation of the impact of the indiscernibility relation on the fuzzy-rough nearest neighbours algorithm}

\author[1]{Henri Bollaert}\corref{cor1}
\ead{henri.bollaert@ugent.be}
\author[1]{Chris Cornelis}
\ead{chris.cornelis@ugent.be}

\affiliation[1]{
    organization={Department of Applied Mathematics, Computer Science and Statistics, Ghent University},
    addressline = {Krijgslaan 281, S9},
    postcode={9000},
    city={Ghent},
    country={Belgium}}
\cortext[cor1]{Corresponding author}

\begin{abstract}
Fuzzy rough sets are well-suited for working with vague, imprecise or uncertain information and have been succesfully applied in real-world classification problems. One of the prominent representatives of this theory is fuzzy-rough nearest neighbours (FRNN), a classification algorithm based on the classical k-nearest neighbours algorithm. The crux of FRNN is the indiscernibility relation, which measures how similar two elements in the data set of interest are. In this paper, we investigate the impact of this indiscernibility relation on the performance of FRNN classification. In addition to relations based on distance functions and kernels, we also explore the effect of distance metric learning on FRNN for the first time. Furthermore, we also introduce an asymmetric, class-specific relation based on the Mahalanobis distance which uses the correlation within each class, and which shows a significant improvement over the regular Mahalanobis distance, but is still beaten by the Manhattan distance. Overall, the Neighbourhood Components Analysis algorithm is found to be the best performer, trading speed for accuracy. 
\end{abstract}


\end{frontmatter}


\section{Introduction}

Fuzzy rough set theory (FRST) is well suited to dealing with imprecise or uncertain information. As such, it has been applied successfully to real life problems in machine learning \cite{Vluymans2015}, multi-criteria decision making \cite{ZHANG202092} and dimensionality reduction \cite{THANGAVEL20091, YUAN2021107353}.
One of the most popular techniques which makes use of FRST is the fuzzy-rough nearest neighbours (FRNN) algorithm for classification problems\cite{Jensen2008,Jensen2010}. To enhance its robustness, the original FRNN algorithm was extended using ordered weighted average (OWA) based fuzzy rough sets \cite{OWAfuzzyroughsets,scalableOWAFRNN}
Indiscernibility relations, which measure how similar two elements of the universe are, play a defining role in FRNN. However, the impact of these relations on the performance of FRNN has not been thoroughly investigated. Therefore, in this paper, we evaluate a broad variety of options for the indiscernibility relation. For our experimental study, we consider relations based on some of the most used distances and other common relations, as well as relations based on kernels. Additionally, we introduce a new relation based on the Mahalanobis distance which aims at data sets where the intra-class distances are not the same across all classes. Additionally, we evaluate the performance of distance metric learning methods (DML) \cite{tut_dml} in fuzzy-rough nearest neighbours for the first time. We perform this evaluation by comparing the balanced accuracy of FRNN with each of the relations on a large corpus of real-life benchmark data sets with different characteristics and by testing for significant differences by mean of statistical analysis.

The remainder of the paper is structured as follows. In Section \ref{sec:prelims}, we recall important notions about fuzzy sets, (OWA-based) fuzzy rough sets and the fuzzy-rough nearest neighbours algorithm. Section \ref{sec:relations} provides a detailed description of the indiscernibility relations used in our experimental study. We describe the setup of our experiments and discuss our results in Section \ref{sec:experiments}. Finally, we summarize our main conclusions and outline directions for future work in Section \ref{sec:conclusions}.

\section{Preliminaries}\label{sec:prelims}

\subsection{Fuzzy sets}

In this paper, \(U\) will indicate a finite, non-empty set, called the \emph{universe}. A \emph{fuzzy set} \cite{fuzzysets} \(A\) in \(U\) is a function \(A \colon U \to [0,1]\). For \(x\) in \(U\), \(A(x)\) is called the membership degree of \(x\) in \(A\).

In fuzzy logic and fuzzy set theory, we use \emph{triangular norms}, \emph{triangular conorms}, \emph{negators} and \emph{implicators }to generalize the classical connectives conjunction (\(\wedge\)) and disjunction (\(\vee\)), negation (\(\neg\)) and implication (\(\rightarrow\)) respectively. A triangular norm or \emph{t-norm} is a function \(\tnorm \colon [0,1]^2 \to [0,1]\) which has \(1\) as neutral element, is commutative and associative and is increasing in both arguments. A triangular conorm or \emph{t-conorm} is a function \(\tconorm \colon [0,1]^2 \to [0,1]\) with neutral element 0, which is commutative and associative and increasing in both arguments. A negator is a function \(\negator \colon [0,1] \to [0,1]\) which is decreasing and for which the following boundary conditions hold: \(\negator(0) = 1\) and \(\negator(1) = 0\). An \emph{implicator} is a function \(\implicator \colon [0,1]^2 \to [0,1]\) which is decreasing in the first argument and increasing in the second one. Additionally, the following boundary conditions should hold: 
\[\implicator(0,0) = \implicator(0,1) = \implicator(1,1) = 1 \text{ and } \implicator(1,0)=0.\]

A (binary) fuzzy relation on a universe \(U\) is simply a fuzzy subset of \(U \times U\). A fuzzy relation \(R\) is reflexive if \(R(x,x) = 1\) for all elements \(x \in U\). It is symmetric if for every pair of elements \(x, y\) in the universe, \(R(x, y) = R(y,x)\). A fuzzy tolerance relation is a reflexive and symmetric fuzzy relation.

\subsection{Ordered Weighted Average based fuzzy rough sets}
In classical rough set theory \cite{roughsets}, we examine a universe divided into equivalence classes by an equivalence relation. We can then approach a concept (i.e., a subset of \(U\)) using two crisp sets: the lower and the upper approximation. The lower approximation contains all elements which are sure to be part of the concept and is defined as the union of the equivalences classes which are subsets of the concept. The upper approximation, which contains all elements which might be a part of the concept, is defined as the union of all equivalence classes which have a non-empty intersection with the concept in question. In fuzzy rough sets, first introduced in \cite{fuzzyroughsets}, we replace the crisp equivalence relation with a fuzzy relation. In turn, the upper and lower approximations also become fuzzy sets. This variant of a fuzzy rough set is still highly susceptible to noise. As such, we use the \(\OWA\)-based fuzzy rough set model introduced in \cite{OWAfuzzyroughsets}. This concept makes use of ordered weighted average (\(\OWA\)) operators. Consider a set \(S = \{s_1, \dots, s_n\}\) with \(s_1 \geq s_2 \geq \dots \geq s_n\) and an \(n\)-dimensional weight vector \(w = (w_1, \dots, w_n)\) for which 
\[\sum_{i=1}^n w_i = 1\]
holds. The result of the \(\OWA\)-operator corresponding to \(w\) on \(S\) is 
\[\OWA_w\langle S \rangle = \sum_{i=1}^n w_i s_i.\]
Now, consider a binary fuzzy relation \(R\) on \(U\) and a fuzzy set \(A\) in \(U\). Furthermore, let \(\tnorm\) be a t-norm, \(\implicator\) an implicator, and \(\underline{w}\) and \(\overline{w}\) two weight vectors.
\begin{itemize}
    \item The \(\OWA\)-based fuzzy lower approximation \(\underline{A}_R^{\implicator, \underline{w}}\) is the fuzzy set
    \[\underline{A}_R^{\implicator, \underline{w}}(x) = \OWA_{\underline{w}}\langle\implicator(R(x, y), A(y)) \mid y \in U\rangle.\]
    \item The \(\OWA\)-based fuzzy upper approximation \(\overline{A}_R^{\tnorm, \overline{w}}\) is the fuzzy set
    \[\overline{A}_R^{\tnorm, \overline{w}}(x) = \OWA_{\overline{w}}\langle\tnorm(R(x, y), A(y)) \mid y \in U\rangle.\]
\end{itemize}
The \emph{\(\OWA\)-based fuzzy rough set (\(\OWA\)-FRS)} for the fuzzy set \(A\), the fuzzy relation \(R\) and the weight vectors \(\underline{w}, \overline{w}\) is the pair \((\underline{A}_R^{\implicator, \underline{w}}, \overline{A}_R^{\tnorm, \overline{w}})\).

\subsection{The fuzzy-rough nearest neighbours algorithm}\label{sec:frnn}

An information system \cite{roughsets} is a pair \((U, \att)\), with \(U\)  a finite, non-empty universe of objects and \(\att\) a finite, non-empty set of attributes which these objects have. Each attribute \(a \in \att\) has an associated domain of possible values \(V_a\) and an associated function \(a \colon U \to V_A\) which maps an object \(x \in U\) to its value \(a(x)\) for the attribute \(a\).
A decision system \((U, \att \cup \{b\})\) with a single decision attribute \(b\) is an information system which makes a distinction between the conditional attributes \(\att\) and the decision attribute \(b\), which is not an element of \(\att\). In this paper, we only consider numerical conditional attributes and categorical decision attributes. An attribute is numerical if its value set is a real interval. A categorical attribute has a finite, unordered value set, of which the elements are called classes.

In a classification problem, the universe of our decision system can be divided into two disjoint sets: the training set \(U_{\train}\) and the test set \(U_{\test}\). Our goal is to predict the true class of each element in the test set given the features of each element in the total universe and the classes of the training set.

We can now apply \(\OWA\)-FRS to classification problems with the \emph{\(\OWA\)-based fuzzy-rough nearest neighbours  (\(\OWA\)-FRNN)\footnote{In the remainder of this text, we will omit \(\OWA\) and simply refer to the algorithm as FRNN.}} algorithm. Let \(R\) be a fuzzy tolerance relation on \(U\), and consider two weight vectors \(\underline{w}\) and \(\overline{w}\) in addition to a t-norm \(\tnorm\) and an implicator \(\implicator\). We construct the \(\OWA\)-FRS \((\underline{C}_R^{\implicator, \underline{w}}, \overline{C}_R^{\tnorm, \overline{w}})\) for each class \(C\) and calculate the membership degree of the new sample \(v\) to its upper and lower approximations. We then assign \(v\) to the class for which the sum of these membership degrees is the highest, i.e., the class identified by
\begin{align*}
    &\arg \max_{C \in V_b} \left( \underline{C}_R^{\implicator + \underline{w}}(v) + \overline{C}_R^{\tnorm, \overline{w}}(v) \right) \\
    = &\arg \max_{C \in V_b} \left( \OWA_{\underline{w}}\langle\implicator(R(v, y), C(y)) \mid y \in U\rangle + \OWA_{\overline{w}}\langle\tnorm(R(v, y), C(y)) \mid y \in U\rangle \right)
\end{align*}
which, on the condition that \(\implicator(x,0) = 1 - x\) holds for all \(x \in [0,1]\), and taking into account that decision classes are crisp, can be simplified to
\[\arg \max_{C \in V_b} \left( \OWA_{\underline{w}}\langle 1 - R(v, y) \mid y \in U \setminus C\rangle + \OWA_{\overline{w}}\langle R(v, y) \mid y \in C\rangle \right).\]

In our implementation of this algorithm, we used \emph{linear} (or \emph{additive}) weights, as these have been found to be the best in general use cases \cite{additive-weights}. Additionally, we will set most of the weights to be zero, thus using only the information of a limited amount of training elements both inside and outside of each class. We will call this amount \(k\). The resulting weights are defined as follows:
\[\overline{w} = \left\langle \frac{2(k+1-i)}{k(k+1)}\right\rangle_{1 \leq i \leq k} , \; \underline{w} = \left\langle \frac{2i}{k(k+1)}\right\rangle_{1 \leq i \leq k},\]
where the remaining components of the vectors are 0. 


\section{Indiscernibility relations}\label{sec:relations}

The main goal of this paper is to determine the effect of the indiscernibility relation on the performance of the \(\OWA\)-FRNN algorithm, and whether distance metric learning makes a significant difference. In this section, we briefly recall the indiscernibility relations under consideration.

\subsection{Indiscernibility relations based on distance functions}\label{sec:classical-relations}

\begin{table}[htp]
    \centering
    \(\begin{array}{l>{\displaystyle}l>{\displaystyle}c}
    \toprule
    \text{name} & f(x,y) & \max_{u,v \in [0,1]^n} f(u,v) \\
    \midrule
    \text{Manhattan} & \sum_{i=1}^n |x_{i} - y_{i}| & n \\[4mm]
    \text{Euclidean} & \sqrt{\sum_{i=1}^n (x_{i} - y_{i})^2} & \sqrt{n}\\[5mm]
    \text{Chebyshev} & \max_{i=1}^n |x_{i} - y_{i}| & 1\\[4mm]
    \text{Canberra} & \sum_{i=1}^n \frac{|x_{i} - y_{i}|}{|x_{i}| + |y_{i}|} & n\\[5mm]
    \text{Mahalanobis} & \sqrt{(x - y)^\top M(x-y)} & - \\[3mm]
    \text{Cosine distance} & 1 - \frac{x^\top y}{x^\top x \cdot y^\top y} & 2 \\[4mm]
    \text{PCC distance} & 1 - \frac{\sum_{i=1}^m (x_{i} - a_i)(y_{i} - a_i)}{\sqrt{\sum_{i=1}^m (x_{i} - a_i)^2 \sum_{i=1}^m (y_{i} - a_i)^2}} & 2 \\
    \bottomrule
    \end{array}\)
    \caption{Distance functions defined on \(\mathbb{R}^n\), for a positive integer \(n\), along with their maximum value on the zero-centered ball with radius 1. The Canberra distance is from \cite{canberra1, canberra2}. In the Mahalanobis distance \cite{mahalanobis}, \(M\) is a positive semi-definite \(n \times n\)-matrix. In the Pearson correlation coefficient (PCC) distance, \(a_i\) is the average of feature \(i\) across \(U\).}
    \label{tab:def-distance-relations}
\end{table}

The first set of relations are defined using distance functions on \(\mathbb{R}^n\), which are \(\mathbb{R}^n \times \mathbb{R}^n \to \mathbb{R}\) functions that are non-negative, reflexive and symmetric. The distance functions which we will consider can be found in Table \ref{tab:def-distance-relations}. We selected these from the most used ones from \cite{EncyclopediaofDistances, improvedhetrogenousdistances}.
A distance function \(d(x,y)\) is converted into a fuzzy indiscernibility relation \(R_d\) by dividing the distance by its maximal value and subtracting it from 1:
\[R_d(x,y) = 1 - \frac{d(x,y)}{\max_{u,v \in U}d(u,v)}\]
This maximal value can be obtained in two ways. For some distances, we can derive the theoretical maximal distance under a certain type of normalisation in a real-valued metric space with a given dimension \(n\). These theoretical values for range normalisation can be found in the third column of Table \ref{tab:def-distance-relations}. However, for Mahalanobis distances, we will explicitly calculate the maximal distance between two elements of the universe. 

We offer some further remarks on Mahalanobis distances. First off, note that the Euclidean distance can be seen as a Mahalanobis distance where \(M\) is the identity matrix. As a more useful candidate for \(M\), we consider the inverse of the covariance matrix of the decision system \cite{mahalanobis}. Note that when there are linearly dependent features in the data set, its covariance matrix is singular and therefore distances cannot be calculated. In addition to the above methods, we propose a class-specific variant of the Mahalanobis distance in the next section.
Second, an often used variant of the Mahalanobis distance is the squared Mahalanobis distance
\[(x - y)^\top M(x-y).\]
This also applies to the Euclidean distance. In our tests, only the distance metric learning methods (DML) discussed in Section \ref{sec:dml} will use a squared Mahalanobis distance.

\subsection{Class-specific Mahalanobis-based relation (CSMBR)}\label{sec:csmbr}

In many real world problems, we discern samples in different ways depending on their class. For example, in the wisconsin data set (see Section \ref{sec:setup}), the intra-class Manhattan distance for one class is 6.5, while it is 27.3 for the other class. In such a data set, there are likely to be samples which are more similar to elements of the other class than to elements of their own class. We might therefore expect a non-symmetric similarity relation which uses the class of its first argument to improve the classification performance of FRNN. Recall that we know the class of at least one argument of any similarity evaluation during the FRNN algorithm, since we only calculate the similarity between a test sample and a training sample.

We apply this idea to the Mahalanobis-based similarity relation. More precisely, consider a decision system \((U, \att \cup \{b\})\), where \(U\) is the disjoint union of a training set \(U_{\train}\) and a test set \(U_{\test}\). For each class \(C \in V_b\), define
\[U_C = \{x \in U_{\train} \mid b(x) = C\}\]
Let \(\hat{U}_C\) be the matrix whose rows consist of the elements of \(U_C\), and \(A_C\) its covariance matrix. We calculate the Mahalanobis-based similarity relation \(R_C\) on \(U_C\) in the usual way:
\[\forall x,y \in U_C: R_C(x,y) = 1 - \frac{\sqrt{(x - y)^T A_C^{-1}(x-y)}}{\max\limits_{u,v \in U_C} \sqrt{(u - v)^T A_C^{-1}(u-v)}}\]
We can then define the CSMBR as
\[\CSMBR \colon U \times U \to [0,1] \colon (x, y) \mapsto R_{b(x)}(x,y)\]

\begin{example}

    \begin{table}[ht]
		\centering
		\begin{tabular}{cllll}
			\toprule
			{}       & \(a_1\)   & \(a_2\)   & \(a_3\)   & \(b\) \\
			\midrule
			\(u\)    & \(0.25\)  & \(0\)     & \(-0.5\)  & \(0\) \\
			\(v\)    & \(0.35\)  & \(-0.25\)  & \(-0.25\) & \(1\) \\
			\midrule
			average  & \(0.25\)  & \(-0.15\)  & \(-0.1\)  & {--}  \\
			\bottomrule
		\end{tabular}
		\caption{Sample values}
		\label{tab:start-example-distances}
	\end{table}

    Consider the following decision system 
	\[(V, \{a_1, a_2, a_3, b\})\]
	and two samples \(u\) and \(v\) from \(V\). Their feature values, as well as the universal feature averages, are given in Table \ref{tab:start-example-distances}.
    Additionally, we assume the following covariance matrices for the subsets of all samples with class 0 and 1 respectively:
	\[A_0 = \begin{pmatrix}
		1 & -0.25 & 0\\
		-0.25 & 1 & 0\\
		0 & 0 & 1\end{pmatrix} \text{ and } A_1 = \begin{pmatrix}
		1 & 0 & 0\\
		0 & 1 & 0.3\\
		0 & 0.3 & 1\end{pmatrix}\]
    Finally, we assume that the maximum distance between two samples of the same class is 1 for both classes.
    This gives us all the information required to calculate the following membership degrees to the class-specific Mahalanobis-based relation (rounded to 2 digits) defined by these covariance matrices.
	\begin{align*}
		&R_\text{cl-mah}(u, v) = R_0(u,v) = 1 - \sqrt{(u - v)^T A_0^{-1}(u - v)} = 0.64 \\
		&R_\text{cl-mah}(v, u) = R_1(v,u) = 1 - \sqrt{(v - u)^T A_1^{-1}(v - u)} = 0.57
	\end{align*}
\end{example}

\subsection{Kernel functions}\label{sec:kernels}

Next, we consider the class of indiscernibility relations based on a kernel function, as introduced in \cite{kernelizedFRS}.
Recall that a kernel is a symmetric and positive semi-definite function \(k \colon \mathbb{R}^d \times \mathbb{R}^d \to \mathbb{R}\).
If the image of a kernel function \(k\) is a subset of the unit interval, we may see it as a fuzzy relation \(R_k\):
\[R_k(x, y) = k(x, y)\]
This relation is symmetric by definition and reflexive if the kernel function is reflexive.
As such, we can use reflexive kernel functions as indiscernibility relations in a decision system. A list of some possible kernel functions along with their partial derivatives in the parameter \(\gamma\), which we will need in Section \ref{sec:kernelfitting}, can be found in Table \ref{tab:def-kernels}. 
Because of the parameter \(\gamma\) in each of these kernel functions, we can fit these functions to the data, as will be explained in the following section.

\begin{table}[htp]
    \centering 
    \(\begin{array}{l>{\displaystyle}l}
        \toprule
        \text{name}        & k(x, y)\\
        \midrule
        \text{Gaussian}    & \exp \left( - \frac{\|x - y\|_2^2}{\gamma}\right)  \\[3mm]
        \text{exponential} & \exp \left( - \frac{\|x - y\|_2}{\gamma}\right)  \\[3mm] 
        \text{rational quadratic} & \frac{\gamma}{\|x - y\|_2^2 + \gamma}  \\[3mm] 
        \text{circular} & \begin{cases} \frac{2}{\pi} \arccos \left( \frac{\|x - y\|_2}{\gamma}\right)  & \\[2mm] \qquad - \frac{2}{\pi} \frac{\|x - y\|_2}{\gamma} \sqrt{1 - \left(\frac{\|x - y\|_2}{\gamma}\right)^2}, &\text{if \(\|x - y\|_2 < \gamma\),} \\ 0, &\text{otherwise.} \end{cases} \\[4mm] 
        \text{spherical} & \begin{cases} 1 - \frac{3}{2} \frac{\|x - y\|_2}{\gamma} + \frac{1}{2} \left(\frac{\|x - y\|_2}{\gamma}\right)^3, &\text{if \(\|x - y\|_2 < \gamma\),} \\ 0, &\text{otherwise.}\end{cases} \\[3mm]
    \bottomrule
    \end{array}\)
    \caption{Kernel functions considered in this paper.}
    \label{tab:def-kernels}
\end{table}

\begin{table}[htp]
    \centering 
    \(\begin{array}{l>{\displaystyle}l}
        \toprule
        \text{name}            & \frac{\partial k_G}{\partial \gamma}(x, y) \\
        \midrule
        \text{Gaussian}    & \frac{\|x - y\|_2^2}{\gamma^2} \exp \left( - \frac{\|x - y\|_2^2}{\gamma}\right) \\[3mm]
        \text{exponential} & \frac{\|x - y\|_2}{\gamma^2} \exp \left( - \frac{\|x - y\|_2}{\gamma}\right) \\[3mm] 
        \text{rational quadratic}  &  \frac{\|x - y\|_2^2}{(\|x - y\|_2^2 + \gamma)^2} \\[3mm] 
        \text{circular} & \begin{cases} 
        \frac{4}{\pi} \frac{ \|x - y\|_2 }{\gamma^2}\sqrt{1 - \frac{\|x - y\|_2^2}{\gamma^2}}, &\text{if \(\|x - y\|_2 < \gamma\),} \\0, &\text{otherwise.} \end{cases} \\[4mm] 
        \text{spherical} & \begin{cases} \frac{3}{2}\frac{\|x - y\|_2}{\gamma^4}(\gamma^2 - \|x - y\|_2^2), &\text{if \(\|x - y\|_2 < \gamma\),} \\ 0, &\text{otherwise.} \end{cases} \\[3mm]
    \bottomrule
    \end{array}\)
    \caption{Derivatives of the kernels}
    \label{tab:def-der-kernels}
\end{table}

\subsection{Distance metric learning for indiscernibility relations}\label{sec:dml}

\emph{Distance metric learning (DML)} \cite{tut_dml} is a domain of machine learning that aims to learn the similarities between different objects, typically with the aim of improving the performance of similarity-based algorithms such as kNN and FRNN. Most of the research is focused on learning a Mahalanobis distance function, due to its ease of parametrization.
We will also evaluate some techniques from the closely related area of indiscernibility learning, where we will learn an indiscernibility relation from the data instead of a distance function.

\subsubsection{Mahalanobis-based distance metric learning}\label{sec:mahdml}

We include three Mahalanobis-based distance metric learning algorithms in the comparison. 
These algorithms use quadratic Mahalanobis distances, since their derivatives are simpler. These derivatives play an important role in the optimisation process of the objective functions of these algorithms, which is done using gradient descent.
As the matrix \(M\), which defines a Mahalanobis distance, is positive semi-definite, there is a matrix \(L\) for which \(M = L^\top L\) holds, such that the distance between two samples given by the Mahalanobis distance w.r.t.\ \(M\) in the original space is the same as the euclidean distance between the images of those samples under the linear transformation defined by \(L\). Therefore, we can define an objective function on the set of all possible linear transformations \(L\) instead of over the set of all positive semi-definite matrices.

We selected three prominent representatives of such algorithms from \cite{tut_dml}. Two of these methods, which are recalled below, have been developed with kNN in mind whereas the final one takes its inspiration from information theory.
\begin{itemize}
    \item Large Margin Nearest Neighbours (LMNN) \textbf{}\cite{LMNN1, LMNN2} aims to increase the accuracy of kNN by looking for a linear transformation to a space where the \(k\) nearest neighbours of a sample belong to the same class and elements of different classes are separated by a large margin.
    \item Neighbourhood components analysis (NCA) \cite{NCA-paper} searches for the linear transformation which minimizes the expected leave-one-out error of kNN on the training set.
    \item Distance metric learning through the maximisation of the Jeffrey Divergence (DMLMJ) \cite{jeffreydml-paper, jeffreydml-thesis} uses techniques from information theory to learn a Mahalanobis pseudo-metric which maximises the difference between two distributions based on the differences between samples of the same class and between samples of different classes. We use the Jeffrey divergence to express this difference. This is a symmetrized version of the Kullback-Leibler divergence, which measures the relative entropy of a query distribution against a reference distribution.
\end{itemize}

\subsubsection{Kernel-based indiscernibility learning}\label{sec:kernelfitting}

A second type of indiscernibility learning we will evaluate is based on the kernel functions from Section \ref{sec:kernels}. As mentioned there, we can fit a kernel to the data by selecting the optimal value for \(\gamma\). We use the gradient descent method \cite{robbins-monro51} to iteratively find the optimal value for \(\gamma\). By using categorical cross entropy and the weights assigned to each class by FRNN, we obtain a differentiable loss function which can be optimised using gradient descent. The required derivatives of each kernel function were already listed in Table \ref{tab:def-kernels}.

\subsubsection{The indiscernibility relation as a hyperparameter}\label{sec:combo}

Even without a parametrized family, we can apply indiscernibility learning by considering the indiscernibility relation as a hyperparameter of the FRNN algorithm and tuning it like any other parameter, e.g.\ with cross-validation on the training set. We applied this idea in an algorithm we called \code{COMBO}. More precisely, we evaluated the performance of a number of indiscernibility relations on the training set using cross-validation and used the best performing relation on the test set.

A short note on the complexity of this algorithm: if we use \(k\) folds to compare \(m\) distances on a data set with \(n\) samples, we need to calculate the distance between \[m \cdot k \cdot \frac{n}{k} \cdot \left(n - \frac{n}{k}\right) = \frac{(k-1)mn^2}{k}\] pairs of samples, since we have to calculate the closest neighbours from the training set for each element in the validation set. This is an increasing function in \(k\), with a smallest value of \(\frac{mn^2}{2}\) for \(k = 2\) and tending to \(m(n-1)n\).


\section{Experiments and results}\label{sec:experiments}

In this section, we will describe our experiments and analyse their results, to study the effects of the indiscernibility relation on the performance of FRNN. After a brief overview of the packages, data sets and performance metrics we used (\Cref{sec:methods}), we move on to the experimental comparisons (\Crefrange{sec:exp:generaldistances}{sec:exp:dml}).

\subsection{Methods}\label{sec:methods}

\subsubsection{Code and packages}\label{sec:code-packages}

All of the experiments and subsequent analysis are done in Python 3.10. For the statistical analysis, we use the \code{scipy.stats} \cite{scipy} and \code{scikit-posthocs} \cite{scikit-posthocs} packages. The implementation of the \(\OWA\)-FRNN algorithm is from the \code{fuzzy-rough-learn} library \cite{lenz20fuzzyroughlearn}. Finally, the implementations of the LMNN, NCA and DMLMJ algorithms are adapted from the PyDML package \cite{pydml}. All other implementations are original, although we use some functions from the \code{numpy} \cite{numpy}, \code{scikit-learn} \cite{scikit-learn} and \code{pandas} \cite{pandaspaper} packages. The entire codebase from this project, including a clone of the \code{fuzzy-rough-learn} library, can be found on GitHub\footnote{\url{https://github.com/henribollaert/frlearn-measures}}.

\subsubsection{Experimental setup}\label{sec:setup}

\begin{table}[!hp]
	\centering
	\begin{tabular}{lrrrr}
		\toprule
		name        & \# num feats &  \#instances & \#classes & IR \\
		\midrule
		aus         & 14 & 690  & 2  & 1.25   \\
		auto        & 15 & 159  & 6  & 16.00  \\
		balance     & 2  & 625  & 3  & 5.88   \\
		banana      & 2  & 5300 & 2  & 1.23   \\
		bands       & 19 & 365  & 2  & 1.70   \\
		bupa        & 6  & 345  & 2  & 1.38   \\
		cleve       & 13 & 297  & 5  & 12.62  \\
		contra      & 9  & 1473 & 3  & 1.89   \\
		crx         & 6  & 653  & 2  & 1.21   \\
		dermatology & 34 & 358  & 6  & 5.55   \\
		ecoli       & 7  & 336  & 8  & 28.60  \\
		german      & 7  & 1000 & 2  & 2.34   \\
		glass       & 9  & 214  & 6  & 8.44   \\
		haberman    & 3  & 306  & 2  & 2.78   \\
		heart       & 13 & 270  & 2  & 1.25   \\
		ion         & 33 & 351  & 2  & 1.79   \\
		mammo       & 5  & 830  & 2  & 1.06   \\
		monk        & 6  & 432  & 2  & 1.12   \\
		movement    & 90 & 360  & 15 & 1.00   \\
		page        & 10 & 5472 & 5  & 175.46 \\
		phoneme     & 5  & 5404 & 2  & 2.41   \\
		pima        & 8  & 768  & 2  & 1.87   \\
		ring        & 20 & 7400 & 2  & 1.02   \\
		saheart     & 8  & 462  & 2  & 1.89   \\
		satimage    & 36 & 6435 & 6  & 2.45   \\
		segment     & 19 & 2310 & 7  & 1.00   \\
		sonar       & 60 & 208  & 2  & 1.14   \\
		spectf      & 44 & 267  & 2  & 38.56  \\
		texture     & 40 & 5500 & 11 & 1.00   \\
		thyroid     & 21 & 7200 & 3  & 40.16  \\
		titanic     & 3  & 2201 & 2  & 2.10   \\
		twonorm     & 20 & 7400 & 2  & 1.00   \\
		vehicle     & 18 & 846  & 4  & 1.10   \\
		vowel       & 13 & 990  & 11 & 1.00   \\
		wdbc        & 30 & 569  & 2  & 1.68   \\
		wine        & 13 & 178  & 3  & 1.48   \\
		red         & 11 & 1599 & 6  & 68.10  \\
		white       & 11 & 4898 & 7  & 439.60 \\
		wisconsin   & 9  & 683  & 2  & 1.86   \\
		yeast       & 8  & 1484 & 10 & 92.60  \\
		\bottomrule
	\end{tabular}
	\caption{Information about the benchmark data sets used for the experiments}
	\label{tab:data sets}
\end{table}

For all of our experiments, we use data sets from the KEEL repository \cite{keeldatarepo}. Some information about these data sets can be found in Table \ref{tab:data sets}. The selection includes a good variety of data sets with different sizes, numbers of attributes and imbalance ratios. Some data sets include categorical features, but as we are only interested in indiscernibility relations defined on numerical attributes, we did not use these attributes in our experiments. We use balanced accuracy as the performance measure.

We use 3 neighbours for both the lower and upper approximations in all experiments. This allows us to look at multiple neighbours while keeping the time complexity down. 
To show that we may restrict ourselves to \(k = 3\), we perform a sensitivity analysis on the number of neighbours for some of the best performing relations in Section \ref{sec:exp:ksens}. As mentioned in Section \ref{sec:frnn}, we used linear weights for the \(\OWA\) weight vectors.

We use the following set-up for the  evaluation of the effect of different indiscernibility relations on the performance of the FRNN algorithm. We partition each data set into ten folds, which we use for ten-fold cross-validation. We use the same partitions for each experiment. For each fold, we first normalise the training data using range normalisation. Then we check whether the indiscernibility relation we want to test is defined on both the test and training sets, which might not be the case, e.g. for the indiscernibility relation based on the Mahalanobis distance if the covariance matrix is not invertible. 
In the case of the distance metric learning methods, we first fit the relation to the data.
Afterwards, we create an instance of the FRNN classifier with this relation on the training set. With all this done, we normalise the test set using the same transformation as the training set, and we determine the predictions of the instantiated classifier for the elements of the test set. These predictions can then be used to calculate the balanced accuracy on that fold.

Following the recommendations of \cite{demsar-howto-comparison}, we use the Wilcoxon signed rank test \cite{pairwise-comp-wilcoxon} to compare the performance of two relations. For multiple comparisons, we apply a two-stage procedure, which starts with a Friedman test \cite{multiple-comp-friedman}. If the result of that test is significant, we continue with a Conover post-hoc pairwise comparison procedure \cite{multiple-comp-conover, conover_test_big_citations} to determine the location of the significant differences.

A cross as an entry in any of the following tables indicates that the indiscernibility relation corresponding to the column of the entry was not defined on any of the folds of the data set corresponding to the row of the entry, a dash indicates that the run time was exceedingly long.

\subsection{Comparison of classical indiscernibility relations}\label{sec:exp:generaldistances}

We start with a comparison of the performance of FRNN with the classical distance-based indiscernibility relations for numerical data from Section \ref{sec:classical-relations}. In particular, we will use this first experiment to compare indiscernibility relations based on:
\begin{itemize}
    \item the Manhattan distance (man),
    \item the Euclidean distance (euc),
    \item the Chebyshev distance (che),
    \item the Canberra distance (can),
    \item the Pearson correlation coefficient (pcc),
    \item the cosine similarity measure (cos),
    \item the Mahalanobis distance with \(M\) equal to the inverse of the covariance matrix (mah).
\end{itemize}

The average balanced accuracy across the ten folds of FRNN using each indiscernibility relation on every data set can be found in Table \ref{tab:distances}.

\begin{table}[!hp]
	\centering
	\begin{tabular}{lrrrrrrr}
		\toprule
            data set      &             man &             euc &             che &             can &             pcc &             cos &             mah \\
		\midrule
		aus           &  \textbf{0.848} &           0.834 &           0.509 &           0.843 &           0.843 &           0.841 &           0.821 \\
		auto          &           0.703 &  \textbf{0.731} &           0.520 &           0.710 &           0.653 &           0.653 &           0.622 \\
		balance       &           0.609 &           0.609 &           0.449 &              -- &  \textbf{0.808} &              -- &              -- \\
		banana        &           0.887 &           0.887 &           0.546 &           0.884 &           0.698 &           0.721 &  \textbf{0.889} \\
		bands         &           0.712 &           0.700 &           0.555 &           0.642 &  \textbf{0.726} &           0.704 &           0.716 \\
		bupa          &           0.647 &           0.635 &           0.626 &           0.622 &           0.653 &           0.625 &  \textbf{0.680} \\
		cleve         &  \textbf{0.332} &           0.304 &           0.210 &           0.296 &           0.311 &           0.315 &           0.295 \\
		contra        &           0.436 &           0.431 &           0.336 &           0.438 &           0.438 &           0.435 &  \textbf{0.450} \\
		crx           &           0.685 &           0.696 &           0.512 &           0.627 &  \textbf{0.723} &           0.683 &           0.721 \\
		dermatology   &  \textbf{0.973} &           0.964 &           0.620 &           0.952 &           0.963 &           0.960 &           0.928 \\
		ecoli         &           0.749 &           0.747 &           0.260 &           0.741 &           0.733 &  \textbf{0.755} &             x \\
		german        &           0.549 &           0.538 &           0.499 &  \textbf{0.555} &           0.538 &           0.545 &           0.549 \\
		glass         &  \textbf{0.697} &           0.672 &           0.672 &           0.682 &           0.691 &           0.662 &           0.674 \\
		haberman      &  \textbf{0.558} &           0.539 &           0.500 &           0.487 &           0.548 &           0.536 &           0.551 \\
		heart         &  \textbf{0.807} &           0.796 &           0.573 &           0.777 &           0.801 &           0.798 &           0.765 \\
		ion           &           0.871 &           0.835 &           0.759 &  \textbf{0.874} &           0.859 &           0.848 &           0.799 \\
		mammo         &           0.802 &           0.802 &           0.500 &           0.791 &  \textbf{0.809} &           0.802 &           0.801 \\
		monk          &  \textbf{0.949} &           0.774 &           0.888 &             -- &           0.850 &             -- &           0.774 \\
		movement      &           0.866 &  \textbf{0.872} &           0.600 &           0.837 &           0.858 &           0.857 &             x \\
		page          &           0.791 &           0.756 &           0.524 &           0.671 &           0.724 &           0.751 &  \textbf{0.818} \\
		phoneme       &  \textbf{0.874} &           0.871 &           0.389 &           0.859 &           0.858 &           0.860 &           0.869 \\
		pima          &           0.671 &           0.674 &           0.504 &           0.661 &           0.683 &  \textbf{0.694} &           0.666 \\
		ring          &           0.699 &  \textbf{0.732} &           0.572 &             -- &           0.690 &             -- &             -- \\
		saheart       &           0.571 &           0.580 &           0.500 &           0.572 &  \textbf{0.598} &           0.571 &           0.565 \\
		satimage      &  \textbf{0.901} &           0.897 &           0.365 &             -- &           0.872 &             -- &             -- \\
		segment       &  \textbf{0.974} &           0.971 &           0.906 &             -- &           0.965 &             -- &             -- \\
		sonar         &           0.849 &           0.851 &           0.667 &           0.834 &  \textbf{0.885} &           0.868 &             x \\
		spectfheart   &           0.605 &           0.565 &           0.500 &  \textbf{0.656} &           0.606 &           0.598 &             x \\
		texture       &           0.988 &           0.990 &           0.185 &           0.980 &  \textbf{0.992} &           0.990 &             x \\
		thyroid       &           0.598 &           0.595 &  \textbf{0.651} &           0.512 &           0.583 &           0.581 &             x \\
		titanic       &           0.532 &           0.532 &  \textbf{0.664} &           0.532 &           0.532 &           0.532 &           0.532 \\
		twonorm       &           0.961 &           0.963 &           0.500 &             -- &  \textbf{0.966} &             -- &           0.886 \\
		vehicle       &           0.709 &           0.722 &           0.595 &           0.705 &           0.721 &           0.718 &  \textbf{0.797} \\
		vowel         &           0.987 &  \textbf{0.990} &           0.940 &           0.965 &  \textbf{0.990} &           0.989 &           0.981 \\
		wdbc          &  \textbf{0.964} &           0.953 &           0.505 &           0.948 &           0.957 &           0.942 &           0.786 \\
		wine          &           0.967 &           0.946 &           0.736 &  \textbf{0.975} &           0.946 &           0.940 &           0.940 \\
		red           &           0.362 &           0.362 &           0.287 &           0.350 &           0.339 &           0.331 &  \textbf{0.373} \\
		white         &  \textbf{0.434} &           0.431 &           0.297 &           0.420 &           0.411 &           0.407 &           0.431 \\
		wisconsin     &           0.965 &           0.967 &           0.500 &           0.959 &  \textbf{0.978} &           0.961 &           0.927 \\
		yeast         &           0.537 &           0.536 &           0.228 &           0.496 &           0.538 &  \textbf{0.547} &           0.524 \\
		\midrule
		mean          &  \textbf{0.733} &           0.724 &           0.520 &           0.702 &           0.726 &           0.706 &           0.704 \\
		\bottomrule
	\end{tabular}
	\caption{Balanced accuracy of FRNN with the distance-based indiscernibility relations on the benchmark data sets.}
	\label{tab:distances}
\end{table}

At first glance, there is a clear separation between the different indiscernibility relations. On average, the relation based on the Manhattan distance seems to be the best performer, with the relations based on the Euclidean distance and Pearson correlation not far behind. Then we encounter the cosine relation, along with the relations based on the Mahalanobis and Canberra distances. The relation based on the Chebyshev distance resulted in by far the worst performance of FRNN. We observe the same ordering of the performances when we look at the Friedman ranks of the relations, which are given in Table \ref{tab:friedman_rank_std}.

\begin{table}[ht]
	\centering
	\begin{tabular}{lr}
		\toprule
		relation & Friedman rank \\
		\midrule
		man      &  \(2.405\) \\
		pcc      &  \(2.881\) \\
		euc      &  \(3.083\) \\
		mah     &  \(3.900\) \\
		cos &  \(4.000\) \\
		can &  \(4.368\) \\
		che      &  \(5.881\) \\
		\bottomrule
	\end{tabular}
	\caption{The Friedman rank of each distance-based indiscernibility relation.}
	\label{tab:friedman_rank_std}
\end{table}

When we look at the maximal difference of each distance to the best performing distance on the same data set in Table \ref{tab:max_dif_std}, we see that the Manhattan based-relation achieves at most \(20\%\) less accurate predictions when compared to the best relation. On average, all but the Chebyshev-based relation are about two to four percent less accurate than the best performing indiscernibility relation. The Chebyshev-based indiscernibility relation causes the FRNN algorithm to be on average more than \(20\%\) less accurate than the best relations and this difference increases to over \(80\%\) on some data sets. The Canberra relation jumps out as the relation which obtains the smallest maximal accuracy loss, at just under \(15\%\).

One may wonder why the indiscernibility relation based on the Chebyshev distance is almost always such a bad performer, while also being the best indiscernibility relation by a large margin on two data sets (thyroid and titanic). At the surface, these data sets are not that different from the ring data set, with around the same number of features (20), samples (around 7000) and classes (2 to 4), in combination with a very low imbalance ratio. However, on the ring data set, FRNN again obtains the lowest balanced accuracy with the Chebyshev-based relation, only barely outperforming random assignment.
The Chebyshev distance is of course very limited, in that it only looks at the largest difference between the features of two samples. As such it is very sensitive to noise and outliers. Therefore, as the statistical analysis will also discover, we do not recommend the Chebyshev distance as a basis for an indiscernibility relation when balanced accuracy is of any concern. 

\begin{table}[ht]
	\centering
	\begin{tabular}{lrr}
		\toprule
		relation &        max &        avg \\
		\midrule
		man      &  \(0.199\) &  \(0.021\) \\
		euc      &  \(0.200\) &  \(0.030\) \\
		che      &  \(0.807\) &  \(0.234\) \\
		can &  \(0.147\) &  \(0.039\) \\
		pcc      &  \(0.190\) &  \(0.028\) \\
		cos &  \(0.168\) &  \(0.034\) \\
		mah   &  \(0.179\) &  \(0.038\) \\
		\bottomrule
	\end{tabular}
	\caption{The maximal and average difference between each relation and the best performing relation.}
	\label{tab:max_dif_std}
\end{table}

Using a Friedman test, we can conclude with a significance level smaller than \(10^{-21}\) that there is a significant difference between the methods. 
Using a Conover post-hoc procedure with Holm corrections, we obtain the adjusted p-values in Table \ref{tab:con-post-hoc-std} for the difference in FRNN's accuracy between each pair of indiscernibility relations. There is a significant difference between almost all pairs, except between the Canberra distance-based relation and the Mahalanobis based relation. As the most important conclusion, we can point out that the performance of FRNN with the indiscernibility relation based on the Manhattan distance is significantly better than its performance with any other distance-based indiscernibility relation.

\begin{table}[htp]
	\centering
	
	\begin{tabular}{lrrrrrr}
		\toprule
		relation &          euc &        che &         can &         pcc &         cos &          mah \\
		\midrule
		man & \(<0.001\) &  \(<0.001\) &  \(<0.001\) &  \(<0.001\) &  \(<0.001\) &  \(<0.001\) \\
		euc &            &  \(<0.001\) &  \(<0.001\) &   \(0.008\) &  \(<0.001\) &  \(<0.001\) \\
		che &            &             &  \(<0.001\) &  \(<0.001\) &  \(<0.001\) &  \(<0.001\) \\
		can &            &             &             &  \(<0.001\) &  \(<0.001\) &   \(0.609\) \\
		pcc &            &             &             &             &  \(<0.001\) &  \(<0.001\) \\
		cos &            &             &             &             &             &  \(<0.001\) \\
		\bottomrule
	\end{tabular}
	\caption{Holm-adjusted p-values obtained with a Conover post-hoc procedure for the standard relations.}
	\label{tab:con-post-hoc-std}
\end{table}

\subsection{Class-specific Mahalanobis-based relation}\label{sec:exp:cl-mah}

\begin{table}[!hp]
	\centering
	\begin{tabular}{lrrr}
		\toprule
		data set        &    man &             mah &       class-mah \\
		\midrule
		aus             &  \textbf{0.848} &           0.821 &           0.827 \\
		auto            &  \textbf{0.703} &           0.622 &           0.516 \\
		banana          &           0.887 &  \textbf{0.889} &           0.887 \\
		bupa            &           0.647 &  \textbf{0.680} &           0.625 \\
		cleve           &  \textbf{0.332} &           0.295 &           0.303 \\
		contra          &           0.436 &           0.450 &  \textbf{0.462} \\
		crx             &           0.685 &  \textbf{0.721} &           0.678 \\
		dermatology     &  \textbf{0.973} &           0.928 &           0.928 \\
		german          &           0.549 &           0.549 &  \textbf{0.565} \\
		glass           &  \textbf{0.697} &           0.674 &           0.529 \\
		haberman        &           0.558 &           0.551 &  \textbf{0.608} \\
		heart           &  \textbf{0.807} &           0.765 &           0.795 \\
		ion             &           0.871 &           0.799 &  \textbf{0.921} \\
		mammo           &  \textbf{0.802} &           0.801 &           0.798 \\
		monk            &  \textbf{0.949} &           0.774 &           0.902 \\
		page            &           0.791 &           0.818 &  \textbf{0.836} \\
		phoneme         &  \textbf{0.874} &           0.869 &           0.831 \\
		pima            &           0.671 &           0.666 &  \textbf{0.687} \\
		saheart         &           0.571 &           0.565 &  \textbf{0.606} \\
		titanic         &  \textbf{0.532} &  \textbf{0.532} &  \textbf{0.532} \\
		twonorm         &  \textbf{0.961} &           0.886 &           0.955 \\
		vehicle         &           0.709 &           0.797 &  \textbf{0.835} \\
		vowel           &  \textbf{0.987} &           0.981 &           0.986 \\
		wdbc            &  \textbf{0.964} &           0.786 &           0.890 \\
		wine            &  \textbf{0.967} &           0.940 &           0.966 \\
		red             &           0.362 &           0.373 &  \textbf{0.374} \\
		wisconsin       &  \textbf{0.965} &           0.927 &           0.910 \\
		yeast           &           0.537 &           0.524 &  \textbf{0.540} \\
		\midrule
		mean            &  \textbf{0.737} &           0.714 &           0.725 \\
		\bottomrule
	\end{tabular}
	\caption{Balanced accuracy of the CSMBR on benchmark data sets, with the normal Mahalanobis relation and the Manhattan relation as a reference. The bolded accuracy is the maximum of the accuracy of FRNN with the standard Mahalanobis relation and with CSMBR.}
	\label{tab:class-specific-mah}
\end{table}

In Table \ref{tab:class-specific-mah}, we evaluate the impact of using class-specific matrices in the relation based on the Mahalanobis distance on the performance of FRNN, as defined in Section \ref{sec:csmbr}.

We will start by covering some details of the implementation of the CSMBR implementation. Ideally, we will calculate a covariance matrix for each class on the set of elements in the training set that belong to that class. However, in practice, this is not always possible. First off, as we saw with the normal Mahalanobis relation, it is possible for a set of samples to have a singular covariance matrix when there are linearly dependent samples in that set. If that is the case for the entire data set, it might also apply to the samples of individual classes. Additionally, in some data sets, the training set might not contain enough samples of a class to be able to calculate the covariance matrix, i.e., when the amount of samples of a class is less than the number of features. Luckily, we can solve this problem by just using the overall covariance matrices in those classes. We need to use this trick in the following data sets: auto, cleve, dermatology, glass, heart, ion, page, vowel, wdbc, wine, red and yeast.

If we turn our attention to Table \ref{tab:class-specific-mah}, we can see that the CSMBR results in on average \(1\%\) more accurate predictions when compared to the classical Mahalanobis distance. A Wilcoxon one-sided test resulted in a p-value of \(0.069\), which suggests that there is weak evidence that this difference is statistically significant. Moreover, the move to CSMBR increases FRNN's accuracy by almost \(10\%\) on some data sets. However, the CSMBR is not always the better of the two relations. For example, on the glass data set, the predictions of FRNN with the normal Mahalanobis relation are almost \(15\%\) more accurate than those with the CSMBR. Additionally, we should note that the Manhattan-based relation is still superior, as can be seen in the table.

\subsection{Kernel-based indiscernibility relations}

\begin{table}[hp]
	\centering
	\begin{tabular}{lrrrrrr}
		\toprule
		data set &             man &           gauss &     exp &         rat &          circle &    sphere \\
		\midrule
		aus        &  \textbf{0.848} &           0.840 &           0.837 &           0.835 &           0.832 &           0.831 \\
		auto        &           0.703 &           0.683 &  \textbf{0.741} &           0.700 &           0.739 &           0.733 \\
		balance           &           0.609 &           0.621 &  \textbf{0.626} &  \textbf{0.626} &  \textbf{0.626} &              -- \\
		banana            &  \textbf{0.887} &  \textbf{0.887} &  \textbf{0.887} &  \textbf{0.887} &  \textbf{0.887} &  \textbf{0.887} \\
		bands             &  \textbf{0.712} &           0.699 &           0.703 &           0.705 &           0.705 &           0.702 \\
		bupa              &  \textbf{0.647} &           0.624 &           0.635 &           0.624 &           0.635 &           0.635 \\
		cleve         &  \textbf{0.332} &           0.307 &           0.305 &           0.310 &           0.317 &           0.317 \\
		contra     &  \textbf{0.436} &           0.433 &           0.434 &           0.434 &           0.430 &           0.431 \\
		crx               &           0.685 &  \textbf{0.704} &           0.695 &  \textbf{0.704} &           0.695 &           0.693 \\
		dermatology       &  \textbf{0.973} &           0.962 &           0.962 &           0.962 &           0.794 &           0.797 \\
		ecoli             &  \textbf{0.749} &           0.745 &           0.747 &           0.745 &           0.747 &           0.747 \\
		german            &  \textbf{0.549} &           0.531 &           0.535 &           0.533 &           0.538 &           0.537 \\
		glass             &  \textbf{0.697} &           0.660 &           0.672 &           0.660 &           0.672 &           0.672 \\
		haberman          &  \textbf{0.558} &           0.532 &           0.537 &           0.532 &           0.539 &           0.539 \\
		heart             &  \textbf{0.807} &           0.797 &           0.792 &           0.797 &           0.763 &           0.762 \\
		ion        &           0.871 &           0.836 &           0.835 &           0.835 &  \textbf{0.929} &           0.926 \\
		mammo      &           0.802 &           0.803 &           0.803 &  \textbf{0.804} &           0.803 &           0.803 \\
		monk              &  \textbf{0.949} &           0.776 &           0.774 &           0.776 &           0.774 &           0.774 \\
		movement   &           0.866 &           0.872 &  \textbf{0.876} &  \textbf{0.876} &           0.809 &           0.809 \\
		page       &  \textbf{0.791} &           0.755 &           0.757 &           0.756 &           0.756 &           0.756 \\
		phoneme           &  \textbf{0.874} &           0.867 &           0.871 &           0.868 &           0.871 &           0.871 \\
		pima              &           0.671 &           0.675 &           0.674 &           0.675 &           0.674 &  \textbf{0.676} \\
		saheart           &           0.571 &           0.580 &  \textbf{0.586} &           0.579 &           0.583 &  \textbf{0.586} \\
		sonar             &  \textbf{0.849} &           0.846 &           0.846 &           0.846 &           0.695 &           0.695 \\
		spectfheart       &  \textbf{0.605} &           0.565 &           0.575 &           0.565 &           0.478 &           0.476 \\
		texture           &           0.988 &  \textbf{0.990} &  \textbf{0.990} &  \textbf{0.990} &  \textbf{0.990} &  \textbf{0.990} \\
		thyroid           &  \textbf{0.598} &           0.585 &           0.597 &           0.585 &           0.594 &           0.594 \\
		titanic           &  \textbf{0.532} &  \textbf{0.532} &  \textbf{0.532} &  \textbf{0.532} &  \textbf{0.532} &  \textbf{0.532} \\
		twonorm           &           0.961 &  \textbf{0.963} &           0.962 &  \textbf{0.963} &  \textbf{0.963} &           0.962 \\
		vehicle           &           0.709 &           0.723 &           0.723 &  \textbf{0.724} &           0.721 &           0.723 \\
		vowel             &           0.987 &           0.989 &  \textbf{0.990} &           0.989 &  \textbf{0.990} &  \textbf{0.990} \\
		wdbc              &  \textbf{0.964} &           0.953 &           0.954 &           0.953 &           0.941 &           0.940 \\
		wine              &  \textbf{0.967} &           0.946 &           0.946 &           0.946 &           0.946 &           0.946 \\
		red   &           0.362 &           0.340 &  \textbf{0.363} &           0.339 &           0.362 &  \textbf{0.363} \\
		white &           0.434 &           0.415 &           0.431 &           0.415 &  \textbf{0.439} &       \textbf{0.439} \\
		wisconsin         &           0.965 &  \textbf{0.967} &           0.965 &  \textbf{0.967} &           0.961 &           0.961 \\
		yeast             &           0.537 &  \textbf{0.543} &           0.530 &           0.541 &           0.531 &           0.530 \\
		\midrule
		mean              &  \textbf{0.733} &           0.718 &           0.721 &           0.718 &           0.710 &           0.712 \\
		\bottomrule
	\end{tabular}
	\caption{Balanced accuracy of FRNN with the kernel-based relations on the benchmark data sets, with the Manhattan relation as a reference}
	\label{tab:kernels}
\end{table}

In this set of experiments, we compare the performance of FRNN with the kernel-based indiscernibility relations from \Cref{sec:kernels}. As a reminder, we compare:
\begin{itemize}
	\item the Gaussian kernel (gauss)
	\item the exponential kernel (exp)
	\item the rational-quadratic kernel (rat)
	\item the circular kernel (circle)
	\item the spherical kernel (sphere)
\end{itemize}

Before using a kernel-based indiscernibility relation, we have to select a value for the parameter \(\gamma\). No guidance is given in \cite{kernelizedFRS, kernelFRNN}. Additionally, the influence of \(\gamma\) depends on the kernel. We can however reason as follows. For small (less than 1) values of \(\gamma\), the kernels lose a lot of information. The circular and spherical kernels set the membership to 0 for all pairs that are further away than \(\gamma\). For the rational quadratic kernel, the influence of the distance between the two samples has a decreased influence. Finally, the smaller \(\gamma\), the faster the Gaussian and exponential kernels tend to 0. For larger values, the circular and spherical kernels become more tolerant and evaluate less similarities as 0. When \(\gamma\) is larger than the maximal Euclidean distance, these kernels are always strictly positive. For the other kernels, the similarity of increasingly distant pairs becomes larger as \(\gamma\) grows.
Therefore, we compromise and set the value of the strictly positive parameter \(\gamma\) equal to 1 for each kernel. 
The average balanced accuracy can be found in Table \ref{tab:kernels}. There are some data sets for which we could not obtain the results of FRNN with kernel-based relations. This is simply because of the time required to run the algorithm on such large data sets. Since the remaining benchmark set is still large enough, we can obtain interesting results.
The kernel-based relations are pretty evenly matched, with the exponential kernel coming out on top with regard to the average balanced accuracy. Furthermore, on most data sets, the performance of FRNN with the kernel-based relation is slightly inferior to that with the Manhattan indiscernibility relation, as we can also see in Table \ref{tab:max_dif_kernels}, where we include the maximum and average difference between each relation and the best performing relation on the same data set. Additionally, we can see that the exponential kernel results in the best performance for the FRNN algorithm when measured using balanced accuracy if we look at the Friedman ranks (Table \ref{tab:friedman_rank_kernels}).

When we perform a Friedman test comparing all of the relations in this experiment, we obtain a significant p-value of less than \(10^{-26}\), and move on to the second stage. The results of the Conover post-hoc tests with Holm-adjusted p-values can be found in Table \ref{tab:con-post-hoc-kernels}. There is a significant difference between the Manhattan relation and all kernel based relations, except for the exponential kernel. However, even in that case, the p-value is quite small (\(<0.10\)). The exponential kernel is in turn significantly different from the Gaussian and spherical kernels. We cannot make a conclusion about any of the other pairs.

\begin{table}[ht]
	\centering
	\begin{tabular}{lrr}
		\toprule
		relation    &        max &        avg \\
		\midrule
		man         &  \(0.058\) &  \(0.005\) \\
		gauss       &  \(0.173\) &  \(0.019\) \\
		exp         &  \(0.175\) &  \(0.015\) \\
		rat         &  \(0.173\) &  \(0.018\) \\
		circle      &  \(0.179\) &  \(0.027\) \\
		sphere      &  \(0.176\) &  \(0.028\) \\
		\bottomrule
	\end{tabular}
	\caption{The maximal, one-but-maximal and average difference of each kernel-based relation and the Manhattan relation to the best performing relation.}
	\label{tab:max_dif_kernels}
\end{table}

Something interesting happens with the monk data set, where there is a difference of more than \(17\%\) accuracy in favour of the Manhattan distance-based relation for all kernels. The circular and spherical relations witness the same drop in accuracy on the dermatology data set, whereas the other three only lose at most \(10\%\) accuracy when compared to the best distance-based relation if we disregard the monk data set. This is peculiar, since the monk data set is a relatively simple and balanced data set with 2 classes and 6 numerical features, whereas the dermatology data set has many more features and is very imbalanced.
On average, the difference between the accuracy of FRNN with a kernel-based relation and with the best performing relation is around \(2\%\), and just \(1.5\%\) for the exponential kernel relation. As such, this relation would be a good candidate for the best overall relation, if we are able to improve its worst performances. In the experiments of Section \ref{sec:exp:kernelfitting}, we will evaluate whether this can be done by tuning the value of \(\gamma\).

\begin{table}[ht]
	\centering
	\begin{tabular}{lr}
		\toprule
		relation    & Friedman rank \\
		\midrule
		man         &  \(2.524\) \\
		exp         &  \(3.203\) \\
		rat         &  \(3.581\) \\
		circle      &  \(3.649\) \\
		sphere      &  \(3.833\) \\
		gauss       &  \(3.946\) \\
		\bottomrule
	\end{tabular}
	\caption{The Friedman rank of the kernel-based and Manhattan-distance based relations.}
	\label{tab:friedman_rank_kernels}
\end{table}

\begin{table}[ht]
	\centering
	\begin{tabular}{lrrrrr}
		\toprule
		relation     &      gauss &         exp &         rat &      circle &       sphere \\
		\midrule
		man          & \(<0.001\) &   \(0.078\) &  \(<0.001\) &  \(<0.001\) &   \(<0.001\) \\
		gauss        &            &   \(0.015\) &   \(0.704\) &   \(0.884\) &    \(1.000\) \\
		exp          &            &             &   \(0.704\) &   \(0.426\) &    \(0.030\) \\
		rat-qua      &            &             &             &   \(1.000\) &    \(0.884\) \\
		circle       &            &             &             &             &    \(0.884\) \\
		\bottomrule
	\end{tabular}
	\caption{Holm-adjusted p-values obtained with a Conover post-hoc procedure for the kernel-based relations and the Manhattan relation.}
	\label{tab:con-post-hoc-kernels}
\end{table}


\subsection{Distance metric learning}\label{sec:exp:dml}

\subsubsection{Mahalanobis-based distance metric learning}\label{sec:exp:mahadml}

\begin{table}[!htp]
	\centering
	\begin{tabular}{lrrrr}
		\toprule
		data set      &             man &             NCA &            LMNN &           DMLMJ \\
		\midrule
		aus    &  \textbf{0.848} &           0.834 &           0.836 &           0.847 \\
		balance       &           0.609 &  \textbf{0.957} &           0.628 &           0.623 \\
		bands         &           0.712 &           0.685 &           0.720 &  \textbf{0.725} \\
		bupa          &           0.647 &           0.635 &           0.626 &  \textbf{0.663} \\
		contra &           0.436 &  \textbf{0.460} &           0.428 &           0.437 \\
		crx           &           0.685 &  \textbf{0.700} &           0.679 &           0.695 \\
		german        &  \textbf{0.549} &           0.541 &           0.535 &           0.547 \\
		haberman      &           0.558 &  \textbf{0.626} &           0.543 &           0.537 \\
		heart         &  \textbf{0.807} &           0.791 &           0.791 &           0.799 \\
		ion    &           0.871 &  \textbf{0.875} &           0.816 &           0.837 \\
		mammo  &           0.802 &  \textbf{0.808} &           0.792 &           0.797 \\
		monk          &           0.949 &  \textbf{1.000} &           0.894 &           0.790 \\
		pima          &           0.671 &           0.681 &  \textbf{0.681} &           0.679 \\
		saheart       &           0.571 &           0.569 &  \textbf{0.581} &           0.572 \\
		sonar         &           0.849 &  \textbf{0.870} &           0.866 &           0.845 \\
		spectfheart   &           0.605 &           0.597 &           0.631 &  \textbf{0.634} \\
		vehicle       &           0.709 &           0.756 &           0.663 &  \textbf{0.764} \\
		vowel         &           0.987 &           0.987 &           0.989 &  \textbf{0.991} \\
		wdbc          &           0.964 &  \textbf{0.969} &           0.958 &           0.962 \\
		wine          &           0.967 &  \textbf{0.984} &           0.974 &           0.977 \\
		wisconsin     &           0.965 &           0.960 &           0.969 &  \textbf{0.969} \\
		\midrule
		mean          &           0.751 &  \textbf{0.775} &           0.743 &           0.747 \\
		\bottomrule
	\end{tabular}
	\caption{Balanced accuracy of FRNN with the DML-based relations on the benchmark data sets, with the Manhattan relation as a reference.}
	\label{tab:dml}
\end{table}

In this experiment, we look at indiscernibility relations which are based on the distance metric learning methods from Section \ref{sec:mahdml}. The three algorithms we will compare in this section are:
\begin{itemize}
	\item Largest Margin Nearest Neighbour (LMNN)
	\item Neighbourhood component analysis (NCA)
	\item Distance metric learning based on the Jeffrey Divergence (DMLMJ)
\end{itemize}
We keep the parameters of the algorithms at the default values of their implementations. The balanced accuracies of FRNN with the relations derived from these three algorithms, along with the performance of FRNN with the Manhattan-based indiscernibility relation can be found in Table \ref{tab:dml}. We omitted some data sets which were too large and resulted in overly long run-times (more than twelve hours).

\begin{table}[!htbp]
	\centering
	\begin{tabular}{lr}
		\toprule
		relation    &        rank\\
		\midrule
		NCA         &  \(2.238\) \\
		DMLMJ       &  \(2.238\) \\
		man         &  \(2.595\) \\
		LMNN        &  \(2.929\) \\
		\bottomrule
	\end{tabular}
	\caption{Friedman ranks of the DML-based relations.}
	\label{tab:friedman-ranks-dml}
\end{table}

We can immediately see that the NCA-based relation performs better than the Manhattan-based relation on average, as we would expect. However, the DMLMJ and LMNN based relations do not. Interestingly enough, the Friedman ranks (Table \ref{tab:friedman-ranks-dml}) tell a slightly different story, where the DMLMJ and NCA relations share the top rank. This is probably because the average accuracy of FRNN with NCA is inflated by its performance on the balance data set, where it is at least \(30\%\) more accurate than any other relation. On the other hand, FRNN with the NCA relation is on average less than \(1\%\) less accurate than the best relation on that data set, and never more than \(4\%\) less accurate (see Table \ref{tab:difs-dml}). Since it is also on average the best performer, it is a top choice as the best relation for any kind of data set. The other relations are also relatively consistent, only achieving \(3 \%\) lower accuracy than the best performer on average.

\begin{table}[!hp]
	\centering
	\begin{tabular}{lrr}
		\toprule
		relation &        max &        avg \\
		\midrule
		man      &  \(0.347\) &  \(0.033\) \\
		NCA      &  \(0.039\) &  \(0.008\) \\
		LMNN     &  \(0.328\) &  \(0.041\) \\
		DMLMJ    &  \(0.334\) &  \(0.037\) \\
		\bottomrule
	\end{tabular}
	\caption{Average and maximal differences between each DML-based relation and the best relation on that data set.}
	\label{tab:difs-dml}
\end{table}

We used the Friedman test to determine whether there is a significant difference and obtained a p-value of less than  \(10^{-7}\), indicating a significant difference among the performances of FRNN with the different DML-based relations. However, a Conover post-hoc test with Holm p-value adjustments did not detect the location of these differences. This is a known phenomenon when doing this kind of statistical analysis \cite{demsar-howto-comparison}, and occurs due to the lack of power of the conservative post-hoc tests. The adjusted p-values are given in Table \ref{tab:dml-conover}. 

\begin{table}[!hp]
	\centering
	\begin{tabular}{lrrr}
		\toprule
		relation    &         NCA &          LMNN &        DMLMJ \\
		\midrule
		man         &   \(1.000\) &     \(1.000\) &    \(1.000\) \\
		NCA         &             &     \(0.532\) &    \(1.000\) \\
		LMNN        &             &               &    \(0.532\) \\
		\bottomrule
	\end{tabular}
	\caption{Holm adjusted p-values for the post-hoc comparison on the DML-based relations.}
	\label{tab:dml-conover}
\end{table}

\subsubsection{Kernel-based indiscernibility learning}\label{sec:exp:kernelfitting}

\begin{table}[!hp]
	\centering
	\begin{tabular}{lrrrrrrr}
		\toprule
		data set          &             exp &      gauss-grad &        exp-grad &        rat-grad &  cir-grad &  sph-grad &      sph-3 \\
		\midrule
		aus        &           0.837 &  \textbf{0.838} &           0.834 &  \textbf{0.838} &     0.556 &     0.541 &           0.834 \\
		auto        &  \textbf{0.741} &           0.673 &           0.731 &           0.673 &     0.353 &     0.358 &           0.731 \\
		banana            &  \textbf{0.887} &  \textbf{0.887} &  \textbf{0.887} &  \textbf{0.887} &     0.500 &     0.500 &  \textbf{0.887} \\
		bands             &  \textbf{0.703} &           0.701 &  \textbf{0.703} &           0.701 &     0.500 &     0.500 &           0.700 \\
		bupa              &  \textbf{0.635} &           0.627 &  \textbf{0.635} &           0.627 &     0.500 &     0.500 &  \textbf{0.635} \\
		cleve         &           0.305 &  \textbf{0.317} &           0.303 &           0.311 &     0.200 &     0.200 &           0.306 \\
		contra     &           0.434 &           0.434 &           0.432 &  \textbf{0.435} &     0.333 &     0.333 &           0.432 \\
		crx               &           0.695 &  \textbf{0.705} &           0.695 &  \textbf{0.705} &     0.500 &     0.500 &           0.696 \\
		dermatology       &           0.962 &  \textbf{0.964} &  \textbf{0.964} &  \textbf{0.964} &     0.167 &     0.167 &  \textbf{0.964} \\
		ecoli             &           0.747 &  \textbf{0.755} &           0.747 &  \textbf{0.755} &     0.174 &     0.174 &           0.747 \\
		german            &           0.535 &           0.532 &  \textbf{0.538} &           0.532 &     0.500 &     0.500 &  \textbf{0.538} \\
		glass             &  \textbf{0.672} &           0.662 &  \textbf{0.672} &           0.662 &     0.170 &     0.170 &  \textbf{0.672} \\
		haberman          &           0.537 &  \textbf{0.541} &  \textbf{0.541} &  \textbf{0.541} &     0.500 &     0.500 &           0.539 \\
		heart             &           0.792 &  \textbf{0.807} &           0.792 &  \textbf{0.807} &     0.500 &     0.500 &           0.792 \\
		ion        &  \textbf{0.835} &           0.828 &  \textbf{0.835} &           0.828 &     0.500 &     0.500 &  \textbf{0.835} \\
		mammo      &  \textbf{0.803} &           0.802 &  \textbf{0.803} &           0.803 &     0.500 &     0.500 &           0.803 \\
		monk              &           0.774 &  \textbf{0.785} &           0.774 &           0.778 &     0.500 &     0.500 &           0.774 \\
		movement  &  \textbf{0.876} &           0.872 &           0.872 &           0.872 &     0.067 &     0.067 &  \textbf{0.876} \\
		page       &  \textbf{0.757} &           0.755 &           0.756 &           0.755 &     0.200 &     0.200 &           0.756 \\
		phoneme           &  \textbf{0.871} &           0.867 & \textbf{0.871} &           0.867 &     0.500 &     0.500 &  \textbf{0.871} \\
		pima              &           0.674 &  \textbf{0.675} &           0.674 &  \textbf{0.675} &     0.500 &     0.500 &           0.674 \\
		saheart           &  \textbf{0.586} &           0.579 &           0.580 &           0.580 &     0.500 &     0.500 &           0.580 \\
		sonar             &           0.846 &  \textbf{0.851} &  \textbf{0.851} &  \textbf{0.851} &     0.500 &     0.500 &           0.846 \\
		spectfheart       &  \textbf{0.575} &           0.567 &  \textbf{0.575} &           0.565 &     0.500 &     0.500 &  \textbf{0.575} \\
		texture           &  \textbf{0.990} &  \textbf{0.990} &  \textbf{0.990} &  \textbf{0.990} &     0.091 &     0.091 &  \textbf{0.990} \\
		thyroid           &  \textbf{0.597} &           0.585 &           0.595 &           0.585 &     0.333 &     0.333 &           0.595 \\
		titanic           &  \textbf{0.532} &  \textbf{0.532} &  \textbf{0.532} &  \textbf{0.532} &     0.500 &     0.500 &  \textbf{0.532} \\
		twonorm           &           0.962 &  \textbf{0.963} &  \textbf{0.963} &  \textbf{0.963} &     0.500 &     0.500 &  \textbf{0.963} \\
		vehicle           &  \textbf{0.723} &  \textbf{0.723} &           0.722 &  \textbf{0.723} &     0.250 &     0.250 &           0.722 \\
		vowel             &  \textbf{0.990} &           0.989 &  \textbf{0.990} &           0.989 &     0.091 &     0.091 &  \textbf{0.990} \\
		wdbc              &  \textbf{0.954} &           0.953 &           0.953 &           0.953 &     0.500 &     0.500 &           0.953 \\
		wine              &  \textbf{0.946} &  \textbf{0.946} &  \textbf{0.946} &  \textbf{0.946} &     0.333 &     0.333 &  \textbf{0.946} \\
		red   &  \textbf{0.363} &           0.340 &           0.362 &           0.340 &     0.167 &     0.167 &           0.362 \\
		white &  \textbf{0.431} &           0.405 &  \textbf{0.431} &           0.413 &     0.155 &     0.155 &  \textbf{0.431} \\
		wisconsin         &           0.965 &  \textbf{0.967} &  \textbf{0.967} &  \textbf{0.967} &     0.500 &     0.500 &  \textbf{0.967} \\
		yeast             &           0.530 &  \textbf{0.544} &           0.536 &  \textbf{0.544} &     0.106 &     0.106 &           0.536 \\
		\midrule
		mean              &  \textbf{0.724} &           0.721 &  \textbf{0.724} &           0.721 &     0.368 &     0.368 &  \textbf{0.724} \\
		\bottomrule
	\end{tabular}
	\caption{Balanced accuracy of FRNN with the kernel-based relations fit with gradient descent on the benchmark data sets.}
	\label{tab:gradient-kernel}
\end{table}

In this set of experiments, we will look at the idea of using gradient descent to learn the optimal value of the strictly positive parameter \(\gamma\) of a kernel-based indiscernibility relation. We apply gradient descent to all of the kernels from Section \ref{sec:kernels}. The balanced accuracy of FRNN with each of these relations can be found in Table \ref{tab:gradient-kernel}, in addition to the accuracy of FRNN with the exponential kernel. For the gradient descent runs, we used a starting \(\gamma\) of 1, a batch size of 10, a learning rate of 0.01, a maximum of 10,000 iterations and a precision of 0.00001. The latter parameter is a lower bound on the difference between two subsequent values from \(\gamma\). If that difference is smaller than the precision threshold, we stop the algorithm, regardless of the number of iterations. If the difference never reaches the threshold, we stop when 10,000 iterations have passed.

The circle and sphere kernels clearly failed at fitting of their parameter using gradient descent. In fact, FRNN is basically reduced to a random classifier when using those relations. This is quite peculiar. As a test, we tried the spherical kernel with different settings for the parameters of gradient descent: an initial \(\gamma\) of 3 and a batch size of 50.  This increased the performance dramatically, and is included in the table in the column sph-3. Furthermore, on most data sets, the algorithm did not use many iterations, often ending in the first twenty, or sometimes even after the first iteration. Moreover, even in the same data set and for the same kernel, one fold might take many hundred iterations and stop at a large value of \(\gamma = 17\), while another might stop immediately. As such, it seems that there is still much room to explore on the optimal values of the parameters of the gradient descent algorithm.

\begin{table}[!ht]
	\centering
	\begin{tabular}{lr}
		\toprule
		relation &          rank\\
		\midrule
		exp-grad    &  \(2.931\) \\
		exp  &  \(2.958\) \\
		sph-3   &  \(2.972\) \\
		rat-grad &  \(3.056\) \\
		gauss-grad  &  \(3.083\) \\
		cir-grad    &  \(6.500\) \\
		sph-grad    &  \(6.500\) \\
		\bottomrule
	\end{tabular}
	\caption{Friedman ranks of the gradient-kernel relations.}
	\label{tab:friedman-ranks-gradkernel}
\end{table}

When we take a look at the average balanced accuracy, we see that using gradient descent does on average not improve the performance of the kernel past the performance of the exponential kernel. However, the Gaussian, rational-quadratic and spherical (with changed parameters) kernels did see an improvement in average balanced accuracy, although it is not statistically significant when compared with a one-sided Wilcoxon test. When ordered according to their Friedman ranks, as in Table \ref{tab:friedman-ranks-gradkernel}, we see that the exponential kernel with gradient descent is ranked first, with the simple exponential kernel not far behind. Then come the Gaussian, rational-quadratic and spherical (with changed parameters) kernels with gradient descent, with the remaining kernels being by far the worst. Disregarding these two kernels, we can see that the kernels are very consistent when compared to each other, as we can surmise from Table \ref{tab:difs-gradkernel}. We already discovered this pattern for kernels without gradient descent optimisation in Table \ref{tab:max_dif_kernels}. 

Unfortunately, we do not see the improvement we hoped for with the monk data set, where all kernels drastically underperformed when compared to the Manhattan relation.

\begin{table}[!ht]
	\centering
	\begin{tabular}{lrr}
		\toprule
		relation &        max &        avg \\
		\midrule
		exp  &  \(0.014\) &  \(0.003\) \\
		gauss-grad  &  \(0.068\) &  \(0.005\) \\
		exp-grad   &  \(0.014\) &  \(0.003\) \\
		rat-grad &  \(0.068\) &  \(0.005\) \\
		sph-3   &  \(0.014\) &  \(0.003\) \\
		\bottomrule
	\end{tabular}
	\caption{Average and maximal differences between each gradient-kernel relation and the best relation on that data set.}
	\label{tab:difs-gradkernel}
\end{table}

Finally, we performed a Friedman test, which was significant (\(p < 10^{-15}\)). The Holm-adjusted p-values from the Conover-post-hoc test can be seen in Table \ref{tab:conover-gradkernel}. We can clearly see that the difference is found between the two underperforming circular and spherical kernels with the standard parameters for gradient descent, while the performance of FRNN with the other kernels are not significantly different.

\begin{table}[!ht]
	\centering
	\begin{tabular}{lrrrrrr}
		\toprule
		relation    &  gauss-grad &    exp-grad &     rat-grad &    cir-grad &    sph-grad &       sph-3 \\
		\midrule
		exp         &   \(1.000\) &   \(1.000\) &    \(1.000\) &  \(<0.001\) &  \(<0.001\) &   \(1.000\) \\
		gauss-grad  &             &   \(1.000\) &    \(1.000\) &  \(<0.001\) &  \(<0.001\) &   \(1.000\) \\
		exp-grad    &             &             &    \(1.000\) &  \(<0.001\) &  \(<0.001\) &   \(1.000\) \\
		rat-grad    &             &             &              &  \(<0.001\) &  \(<0.001\) &   \(1.000\) \\
		cir-grad    &             &             &              &             &   \(1.000\) &  \(<0.001\) \\
		sph-grad    &             &             &              &             &             &  \(<0.001\) \\
		\bottomrule
	\end{tabular}
	\caption{Holm adjusted p-values for the post-hoc comparison on the gradient-kernel relations.}
	\label{tab:conover-gradkernel}
\end{table}

\subsubsection{COMBO}

\begin{table}[!hp]
	\centering
	\begin{tabular}{lrrr}
		\toprule
		data set          &             man &           combo &      best \\
		\midrule
		aus        &  \textbf{0.848} &           0.844 &  0.848 \\
		auto        &           0.703 &  \textbf{0.710} &  0.731 \\
		banana            &  \textbf{0.887} &           0.886 &  0.889 \\
		bands             &  \textbf{0.712} &           0.711 &  0.726 \\
		bupa              &           0.647 &  \textbf{0.658} &  0.680 \\
		cleve         &  \textbf{0.332} &           0.298 &  0.332 \\
		contra     &           0.436 &  \textbf{0.450} &  0.450 \\
		crx               &           0.685 &  \textbf{0.717} &  0.723 \\
		dermatology       &           0.973 &  \textbf{0.977} &  0.973 \\
		ecoli             &  \textbf{0.749} &           0.733 &  0.755 \\
		german            &  \textbf{0.549} &           0.539 &  0.555 \\
		glass             &           0.697 &  \textbf{0.711} &  0.697 \\
		haberman          &  \textbf{0.558} &           0.522 &  0.558 \\
		heart             &  \textbf{0.807} &           0.804 &  0.807 \\
		ion        &  \textbf{0.871} &           0.792 &  0.874 \\
		mammo      &  \textbf{0.802} &           0.792 &  0.809 \\
		movement  &  \textbf{0.866} &  \textbf{0.866} &  0.872 \\
		page       &           0.791 &  \textbf{0.808} &  0.818 \\
		phoneme           &  \textbf{0.874} &  \textbf{0.874} &  0.874 \\
		pima              &           0.671 &  \textbf{0.688} &  0.694 \\
		saheart           &           0.571 &  \textbf{0.601} &  0.598 \\
		sonar             &  \textbf{0.849} &           0.833 &  0.885 \\
		spectfheart       &           0.605 &  \textbf{0.651} &  0.656 \\
		texture           &  \textbf{0.988} &  \textbf{0.988} &  0.992 \\
		thyroid           &           0.598 &  \textbf{0.651} &  0.651 \\
		titanic           &  \textbf{0.532} &  \textbf{0.532} &  0.664 \\
		vehicle           &           0.709 &  \textbf{0.797} &  0.797 \\
		vowel             &  \textbf{0.987} &           0.981 &  0.990 \\
		wdbc              &  \textbf{0.964} &           0.951 &  0.964 \\
		wine              &  \textbf{0.967} &           0.950 &  0.975 \\
		red   &           0.362 &  \textbf{0.374} &  0.373 \\
		white &           0.434 &  \textbf{0.497} &  0.434 \\
		wisconsin         &           0.965 &  \textbf{0.978} &  0.978 \\
		yeast             &           0.537 &  \textbf{0.555} &  0.547 \\
		\midrule
		mean              &           0.721 &  \textbf{0.727} &  0.740 \\
		\bottomrule
	\end{tabular}
	\caption{Balanced accuracy of FRNN on the benchmark data sets with the Manhattan relation, the COMBO relation and the best distance-based relation.}
	\label{tab:comboaccuracy}
\end{table}

Given that on some data sets, using the indiscernibility relation based on the Manhattan distance can result in FRNN making predictions that are almost \(20\%\) lower than when we use the best relation on that data set, experimentation with the \code{COMBO} algorithm from Section \ref{sec:combo} is very much worthwhile. For this experiment, we ran the algorithm with 5 folds and gave it the choice of the seven indiscernibility relations from Section \ref{sec:exp:generaldistances}, and kept the other parameters of the FRNN algorithm the same. In Table \ref{tab:comboaccuracy}, we collected the balanced accuracy of FRNN with the indiscernibility relation based on the Manhattan distance, with the best relation on that data set, and with the relation chosen by the \code{COMBO} algorithm. The bolded score indicates the highest balanced accuracy FRNN obtained with the Manhattan relation or the \code{COMBO} relation. The run time of this algorithm was too long on the following data sets: balance, monk, ring, satimage, segment, twonorm. Their results are therefore omitted from the table. 

From these results, we can see that, on average, the \code{COMBO} algorithm selects a better performing algorithm than the Manhattan relation, with an average increase of \(0.6\%\) accuracy. While we know from our first set of experiments that with the Manhattan relation, FRNN can obtain \(20\%\) less accurate results than with the best distance-based relation on that same data set, the \code{COMBO} algorithm can reduce this maximal difference in accuracy to around \(13\%\), with an average difference below \(2\%\). However, the algorithm does not always succeed at finding the best relation, as we can see when we compare its performance to the best row in Table \ref{tab:comboaccuracy}. Interestingly enough, using a combination of different relations for different folds of a data set, the relations selected by \code{COMBO} outperform the best distance-based relation on five data sets (dermatology, glass, saheart, red and white).

Finally, we performed a Wilcoxon one-sided test to determine whether the \code{COMBO} algorithm improves the predictive accuracy of FRNN significantly when compared to the Manhattan relation. With a p-value of \(0.14\), we were not able to make that conclusion, but we still have some evidence towards this hypothesis.

In conclusion, we would recommend this algorithm, since it is also relatively efficient for smaller values of \(k\), as explained in Section \ref{sec:combo}.

\subsection{Analysis of the impact of the number of neighbours}\label{sec:exp:ksens}

\begin{table}[ht]
    \centering
\begin{tabular}{lllllll}
\toprule
\(k\) &          man &    euc &    che &    can &    cos &    mah \\
\midrule
1  &  \textbf{2.379} &  2.439 &  5.485 &  3.576 &  3.182 &  3.481 \\
3  &  \textbf{2.061} &  2.788 &  5.394 &  3.788 &  3.273 &  3.185 \\
5  &  \textbf{2.242} &  2.652 &  5.364 &  3.697 &  3.500 &  3.000 \\
10 &  \textbf{2.439} &  2.545 &  5.394 &  3.621 &  3.788 &  2.593 \\
15 &  \textbf{2.364} &  2.697 &  5.485 &  3.303 &  3.712 &  2.870 \\
20 &  \textbf{2.348} &  2.818 &  5.303 &  3.439 &  3.576 &  2.963 \\
\bottomrule
\end{tabular}
    \caption{The Friedman ranks for classical relations with \(k \in \{1,3,5,10,15,20\}\).}
    \label{tab:friedman}
\end{table}

We perform a sensitivity analysis on the number of neighbours on the following subset of indiscernibility relations:
\begin{itemize}
    \item the Manhattan distance (man),
    \item the Euclidean distance (euc),
    \item the Chebyshev distance (che),
    \item the Canberra distance (can),
    \item the cosine similarity measure (cos),
    \item the Mahalanobis distance (mah),
\end{itemize}
We evaluate the following numbers of neighbours:
\[\{1,3,5,10,15,20\}\]
and give the balanced accuracy of FRNN with all relations on a subset of data sets in Table \ref{tab:ap:k1} (\(k=1\)), Table \ref{tab:ap:k5} (\(k=5\)), Table \ref{tab:ap:k10} (\(k=10\)), Table \ref{tab:ap:k15} (\(k=15\)) and Table \ref{tab:ap:k20} (\(k=20\)), which can be found in the \hyperlink{app:k-sens}{Appendix}. 
We left out some larger data sets (balance, marketing, ring, satimage, segment, spambase, twonorm, saheart and monk) and those where the Mahalanobis distance was undefined (ecoli, movement, sonar, spectfheart, texture and thyroid).
The results for \(k = 3\) can be found in Section \ref{sec:experiments}. 

We calculate the Friedman ranks of each relation for all values of \(k\), and give the results in Table \ref{tab:friedman}. We can conclude that the general ordering of the indiscernibility relations for \(k = 3\) is retained for different values of \(k\). For all values under consideration, FRNN performs the best with the Manhattan distance, followed by the Euclidean distance. The Chebyshev distance results in the worst performance every time. There are some changes in the middle of the pack distances. The Mahalanobis distance is the third best performer in all cases except when considering only one neighbour, and the cosine distance loses its advantage as the number of neighbours increases. However, the cosine and Canberra-based relations always obtain very similar Friedman ranks.

We also evaluate the performance of FRNN with NCA for the same numbers of neighbours. The results can be found in Table \ref{tab:ap:ncasens}. When we compare these to the results of FRNN with the Manhattan distance for the same \(k\) on the same data set (Table \ref{tab:ap:mansens}, we can see that NCA consistently outperforms the Manhattan distance, and that its winning margin grows as \(k\) increases.

We can conclude that the results of the experiments of the previous sections can be extended to different numbers of neighbours.

\section{Conclusions and future work}\label{sec:conclusions}

As can be seen from the results of our experiments, the choice of indiscernibility relation can have a massive impact on the predictive performance of the FRNN algorithm. In most cases, the distance-based relations, with the notable exception of the Chebyshev distance, are valid options, with the Manhattan distance being the relation that seems to offer FRNN the best balanced accuracy. The class-specific Mahalanobis-based relation performs better than its classical counterpart, but does not reach the level of accuracy of the Manhattan relation. If efficiency is not the main focus, using the \code{COMBO} algorithm is a good option as using it sees FRNN be more consistently accurate. Kernel-based relations are among the better options as well, with most kernels resulting in predictions that are just a tiny bit less accurate than the best distance-based relations. For kernels, it is the exponential kernel that performs the best, though not as well as the Manhattan-distance based relation. Using gradient descent to tune the parameter of these kernels is possible, but requires finessing with the parameters of the algorithm, as wrong settings might result in a decrease in performance.
Finally, the DML methods show great promise. NCA in particular offers a significant increase in predictive accuracy of the FRNN classifier, however, this comes at the cost of a much longer training time.


For researchers with a focus on applied machine learning, further tuning of the parameters of DML algorithms such as LMNN, NCA, and DMLMJ and of kernel-based indiscernibility learning may result in improved performance.

Many directions for future work are possible.
For example, we could apply gradient descent to other parts of the \(\OWA\)-FRNN classifier. In particular, we could use it to find the optimal values for the weight vectors that are used for the \(\OWA\)-based upper and lower approximators of each class, perhaps even in a class-specific way.


Moreover, we should explore the impact of dimensionality reduction on FRNN, in particular in the context of Mahalanobis-based DML methods, where it is easily obtained by using a matrix which is not of full rank.

Additionally, the concept of class-specific relations could be applied to the DML methods, where we learn a different Mahalanobis matrix for each class. However, this would dramatically increase the time and space complexity of the algorithm, and as we have seen with the class-specific gradients, the result might not necessarily improve accuracy.

Finally, the concept of class-specific relations could be extended to ``location-specific'' relations, where we use a relation that is built from fuzzy tolerance relations defined on local subsets of the training set. Of course, such relations would need a large enough data set to be practical, but the CSMBR has shown that aligning our relation to properties of the data set can be advantageous.

\section*{Acknowledgments}

The research reported in this paper was conducted with the financial support of the
Odysseus programme of the Research Foundation – Flanders (FWO). The grant number is G0H9118N. Henri Bollaert would like to thank the Special Research Fund of Ghent University (BOF-UGent) for funding his research.

\newpage

\appendix
\setcounter{table}{0}
\renewcommand\thetable{\roman{table}}
\section*{Appendix: Tables for the analysis of the impact of the number of neighbours}\hypertarget{app:k-sens}{}

\begin{table}[h]
    \centering
    \begin{tabular}{lrrrrrr}
\toprule
data set &             man &             euc &             che &             can &             cos &             mah \\
\midrule
aust        &           0.816 &           0.814 &           0.508 &           0.814 &  \textbf{0.816} &           0.792 \\
auto        &  \textbf{0.774} &           0.738 &           0.548 &           0.768 &           0.738 &           0.631 \\
banana            &           0.873 &           0.873 &           0.583 &           0.870 &           0.680 &  \textbf{0.874} \\
bands             &  \textbf{0.711} &           0.698 &           0.556 &           0.647 &           0.693 &           0.686 \\
bupa              &           0.613 &           0.637 &           0.611 &           0.636 &           0.586 &  \textbf{0.639} \\
cleve         &  \textbf{0.338} &           0.324 &           0.222 &           0.290 &           0.310 &           0.303 \\
contra     &           0.414 &           0.420 &           0.325 &  \textbf{0.430} &           0.416 &           0.426 \\
crx               &           0.645 &           0.650 &           0.493 &           0.606 &           0.628 &  \textbf{0.684} \\
dermatology       &  \textbf{0.955} &           0.949 &           0.641 &           0.925 &           0.944 &           0.922 \\
german            &           0.570 &           0.544 &           0.496 &  \textbf{0.585} &           0.542 &           0.548 \\
glass             &  \textbf{0.711} &           0.704 &           0.652 &           0.696 &           0.703 &           0.642 \\
haberman          &           0.536 &           0.539 &           0.491 &           0.538 &           0.535 &  \textbf{0.564} \\
heart             &  \textbf{0.781} &           0.761 &           0.557 &           0.740 &           0.759 &           0.758 \\
ion        &  \textbf{0.878} &           0.833 &           0.723 &           0.865 &           0.849 &           0.814 \\
mammo      &  \textbf{0.746} &           0.744 &           0.508 &           0.737 &           0.742 &           0.744 \\
page       &           0.761 &           0.731 &           0.457 &           0.693 &           0.723 &  \textbf{0.784} \\
phoneme           &  \textbf{0.874} &           0.873 &           0.457 &           0.855 &           0.862 &           0.866 \\
pima              &           0.656 &  \textbf{0.664} &           0.503 &           0.664 &           0.657 &           0.655 \\
titanic           &           0.521 &           0.521 &  \textbf{0.603} &           0.521 &           0.521 &           0.521 \\
vehicle           &           0.694 &           0.706 &           0.577 &           0.708 &           0.699 &  \textbf{0.785} \\
vowel             &           0.994 &  \textbf{0.995} &           0.946 &           0.969 &           0.994 &           0.986 \\
wdbc              &           0.951 &  \textbf{0.951} &           0.521 &           0.944 &           0.936 &           0.775 \\
wine              &           0.958 &           0.953 &           0.729 &  \textbf{0.960} &           0.946 &           0.919 \\
red   &  \textbf{0.362} &           0.357 &           0.329 &           0.340 &           0.359 &           0.357 \\
white &           0.450 &  \textbf{0.454} &           0.303 &           0.427 &           0.453 &           0.444 \\
wisconsin         &           0.956 &           0.951 &           0.500 &  \textbf{0.957} &           0.956 &           0.929 \\
yeast             &           0.518 &           0.510 &           0.229 &           0.471 &  \textbf{0.519} &           0.502 \\
\midrule
mean              &  \textbf{0.706} &           0.700 &           0.521 &           0.691 &           0.688 &           0.687 \\
\bottomrule
\end{tabular}

    \caption{Balanced accuracy of FRNN on the benchmark data sets with the classical relations, \(k = 1\).}
    \label{tab:ap:k1}
\end{table}

\begin{table}[hp]
    \centering
    \begin{tabular}{lrrrrrr}
\toprule
data set &             man &             euc &             che &             can &             cos &             mah \\
\midrule
aust        &           0.848 &           0.849 &           0.510 &  \textbf{0.858} &           0.846 &           0.830 \\
auto        &           0.672 &  \textbf{0.691} &           0.516 &           0.667 &           0.593 &           0.617 \\
banana            &  \textbf{0.894} &           0.891 &           0.535 &           0.889 &           0.733 &           0.893 \\
bands             &           0.695 &           0.699 &           0.567 &           0.654 &           0.704 &  \textbf{0.710} \\
bupa              &           0.656 &           0.648 &           0.634 &           0.625 &           0.614 &  \textbf{0.675} \\
cleve         &           0.310 &  \textbf{0.325} &           0.216 &           0.301 &           0.302 &           0.294 \\
contra            &           0.454 &           0.444 &           0.336 &           0.444 &           0.447 &  \textbf{0.462} \\
crx               &           0.700 &           0.716 &           0.508 &           0.641 &           0.696 &  \textbf{0.730} \\
dermatology       &  \textbf{0.977} &           0.969 &           0.696 &           0.956 &           0.966 &           0.932 \\
german            &           0.548 &           0.527 &           0.502 &  \textbf{0.557} &           0.530 &           0.543 \\
glass             &  \textbf{0.683} &           0.665 &           0.676 &           0.617 &           0.653 &           0.656 \\
haberman          &           0.530 &           0.527 &           0.500 &           0.502 &           0.511 &  \textbf{0.534} \\
heart             &  \textbf{0.814} &           0.796 &           0.562 &           0.802 &           0.791 &           0.798 \\
ion               &           0.859 &           0.814 &           0.768 &  \textbf{0.877} &           0.820 &           0.781 \\
mammo             &  \textbf{0.812} &           0.809 &           0.501 &           0.797 &           0.811 &           0.803 \\
page              &           0.776 &           0.733 &           0.555 &           0.647 &           0.715 &  \textbf{0.803} \\
phoneme           &  \textbf{0.873} &           0.872 &           0.353 &           0.857 &           0.852 &           0.866 \\
pima              &           0.674 &           0.681 &           0.504 &           0.667 &  \textbf{0.706} &           0.679 \\
titanic           &           0.631 &           0.631 &  \textbf{0.664} &           0.631 &           0.631 &           0.631 \\
vehicle           &           0.714 &           0.723 &           0.616 &           0.705 &           0.714 &  \textbf{0.800} \\
vowel             &  \textbf{0.985} &           0.983 &           0.927 &           0.957 &           0.977 &           0.970 \\
wdbc              &  \textbf{0.966} &           0.959 &           0.500 &           0.947 &           0.942 &           0.761 \\
wine              &           0.967 &           0.961 &           0.746 &  \textbf{0.975} &           0.945 &           0.956 \\
red   &           0.340 &           0.341 &           0.281 &           0.345 &           0.327 &  \textbf{0.353} \\
white &           0.420 &  \textbf{0.422} &           0.278 &           0.400 &           0.368 &           0.421 \\
wisconsin         &  \textbf{0.967} &           0.966 &           0.500 &           0.955 &           0.956 &           0.933 \\
yeast             &           0.547 &           0.554 &           0.203 &           0.507 &  \textbf{0.566} &           0.552 \\
\midrule
mean              &  \textbf{0.715} &           0.711 &           0.524 &           0.696 &           0.693 &           0.703 \\
\bottomrule
\end{tabular}

    \caption{Balanced accuracy of FRNN on the benchmark data sets with the classical relations, \(k = 5\).}
    \label{tab:ap:k5}
\end{table}

\begin{table}[hp]
    \centering
    \begin{tabular}{lrrrrrr}
\toprule
data set &             man &             euc &             che &             can &             cos &             mah \\
\midrule
aust        &           0.854 &           0.844 &           0.503 &  \textbf{0.860} &           0.851 &           0.839 \\
auto        &           0.624 &           0.592 &           0.467 &  \textbf{0.670} &           0.551 &           0.614 \\
banana            &  \textbf{0.897} &           0.895 &           0.535 &           0.894 &           0.751 &           0.896 \\
bands             &           0.662 &           0.679 &           0.595 &           0.643 &           0.679 &  \textbf{0.758} \\
bupa              &           0.672 &           0.657 &           0.615 &           0.619 &           0.636 &  \textbf{0.688} \\
cleve         &           0.294 &           0.301 &           0.203 &           0.268 &           0.280 &  \textbf{0.301} \\
contra     &           0.462 &           0.448 &           0.350 &           0.462 &           0.453 &  \textbf{0.468} \\
crx               &           0.695 &           0.725 &           0.534 &           0.656 &           0.697 &  \textbf{0.727} \\
dermatology       &           0.965 &  \textbf{0.967} &           0.685 &           0.963 &           0.965 &           0.932 \\
german            &           0.527 &           0.518 &           0.514 &  \textbf{0.534} &           0.516 &           0.530 \\
glass             &  \textbf{0.640} &           0.613 &           0.593 &           0.512 &           0.567 &           0.615 \\
haberman          &           0.529 &           0.534 &           0.500 &           0.505 &           0.517 &  \textbf{0.550} \\
heart             &           0.808 &           0.790 &           0.559 &  \textbf{0.808} &           0.793 &           0.804 \\
ion        &           0.855 &           0.800 &           0.791 &  \textbf{0.867} &           0.789 &           0.760 \\
mammo      &           0.816 &  \textbf{0.819} &           0.501 &           0.810 &           0.803 &           0.812 \\
page       &           0.697 &           0.651 &           0.591 &           0.571 &           0.640 &  \textbf{0.769} \\
phoneme           &  \textbf{0.861} &           0.861 &           0.351 &           0.848 &           0.841 &           0.860 \\
pima              &           0.676 &           0.692 &           0.500 &           0.664 &  \textbf{0.710} &           0.671 \\
titanic           &           0.631 &           0.631 &  \textbf{0.664} &           0.631 &           0.631 &           0.631 \\
vehicle           &           0.713 &           0.723 &           0.608 &           0.706 &           0.727 &  \textbf{0.808} \\
vowel             &           0.959 &  \textbf{0.967} &           0.893 &           0.928 &           0.927 &           0.945 \\
wdbc              &           0.962 &  \textbf{0.966} &           0.500 &           0.946 &           0.943 &           0.735 \\
wine              &           0.967 &  \textbf{0.970} &           0.769 &           0.965 &           0.949 &           0.955 \\
red   &           0.331 &           0.336 &           0.275 &           0.315 &           0.311 &  \textbf{0.339} \\
white &           0.366 &           0.358 &           0.264 &           0.350 &           0.327 &  \textbf{0.384} \\
wisconsin         &           0.966 &  \textbf{0.973} &           0.500 &           0.951 &           0.957 &           0.933 \\
yeast             &           0.552 &           0.560 &           0.184 &           0.498 &           0.557 &  \textbf{0.562} \\
\midrule
mean              &  \textbf{0.703} &           0.699 &           0.520 &           0.683 &           0.680 &           0.699 \\
\bottomrule
\end{tabular}

    \caption{Balanced accuracy of FRNN on the benchmark data sets with the classical relations, \(k = 10\).}
    \label{tab:ap:k10}
\end{table}

\begin{table}[hp]
    \centering
    \begin{tabular}{lrrrrrr}
\toprule
data set &             man &             euc &    che &             can &             cos &             mah \\
\midrule
aust        &           0.859 &           0.850 &  0.502 &  \textbf{0.869} &           0.867 &           0.838 \\
auto        &           0.600 &           0.571 &  0.414 &  \textbf{0.667} &           0.503 &           0.489 \\
banana            &  \textbf{0.899} &           0.898 &  0.534 &           0.896 &           0.752 &           0.898 \\
bands             &           0.640 &           0.671 &  0.602 &           0.635 &  \textbf{0.673} &           0.669 \\
bupa              &           0.670 &           0.633 &  0.608 &           0.627 &           0.632 &  \textbf{0.707} \\
cleve         &           0.267 &  \textbf{0.305} &  0.208 &           0.270 &           0.293 &           0.266 \\
contra     &           0.468 &           0.456 &  0.353 &           0.471 &           0.459 &  \textbf{0.476} \\
crx               &           0.688 &           0.715 &  0.538 &           0.656 &           0.696 &  \textbf{0.727} \\
dermatology       &  \textbf{0.968} &           0.964 &  0.661 &           0.957 &           0.959 &           0.939 \\
german            &           0.523 &           0.518 &  0.510 &  \textbf{0.529} &           0.522 &           0.523 \\
glass             &  \textbf{0.623} &           0.613 &  0.575 &           0.505 &           0.526 &           0.517 \\
haberman          &           0.550 &           0.564 &  0.500 &           0.502 &           0.518 &  \textbf{0.570} \\
heart             &           0.806 &           0.798 &  0.547 &  \textbf{0.812} &           0.798 &           0.799 \\
ion        &           0.851 &           0.794 &  0.829 &  \textbf{0.855} &           0.765 &           0.687 \\
mammo      &           0.816 &           0.809 &  0.501 &           0.819 &           0.794 &  \textbf{0.819} \\
page       &           0.646 &           0.634 &  0.600 &           0.553 &           0.608 &  \textbf{0.678} \\
phoneme           &  \textbf{0.853} &           0.852 &  0.373 &           0.841 &           0.826 &           0.849 \\
pima              &           0.691 &           0.699 &  0.500 &           0.664 &  \textbf{0.708} &           0.687 \\
titanic           &           0.681 &           0.681 &  0.664 &  \textbf{0.682} &           0.681 &  \textbf{0.682} \\
vehicle           &           0.709 &           0.707 &  0.608 &           0.708 &           0.718 &  \textbf{0.804} \\
vowel             &  \textbf{0.939} &           0.938 &  0.848 &           0.910 &           0.869 &           0.927 \\
wdbc              &           0.964 &  \textbf{0.967} &  0.500 &           0.939 &           0.937 &           0.725 \\
wine              &  \textbf{0.977} &           0.970 &  0.774 &           0.965 &           0.959 &           0.965 \\
red   &           0.312 &           0.321 &  0.272 &           0.307 &           0.299 &  \textbf{0.333} \\
white &           0.343 &           0.339 &  0.250 &           0.326 &           0.295 &  \textbf{0.355} \\
wisconsin         &           0.962 &  \textbf{0.967} &  0.500 &           0.946 &           0.953 &           0.922 \\
yeast             &           0.552 &           0.556 &  0.182 &           0.488 &           0.554 &  \textbf{0.566} \\
\midrule
mean              &  \textbf{0.698} &           0.696 &  0.517 &           0.681 &           0.673 &           0.682 \\
\bottomrule
\end{tabular}

    \caption{Balanced accuracy of FRNN on the benchmark data sets with the classical relations, \(k = 15\).}
    \label{tab:ap:k15}
\end{table}

\begin{table}[hp]
    \centering
    \begin{tabular}{lrrrrrr}
\toprule
data set &             man &             euc &             che &             can &             cos &             mah \\
\midrule
aust        &           0.863 &           0.855 &           0.502 &  \textbf{0.873} &           0.862 &           0.839 \\
auto        &           0.551 &           0.525 &           0.389 &  \textbf{0.670} &           0.499 &           0.456 \\
banana            &  \textbf{0.902} &           0.900 &           0.533 &           0.895 &           0.751 &           0.900 \\
bands             &           0.628 &  \textbf{0.665} &           0.587 &           0.623 &           0.662 &           0.557 \\
bupa              &           0.664 &           0.632 &           0.601 &           0.633 &           0.643 &  \textbf{0.691} \\
cleve         &           0.281 &           0.270 &           0.200 &  \textbf{0.282} &           0.256 &           0.255 \\
contra     &           0.468 &           0.466 &           0.354 &           0.475 &           0.456 &  \textbf{0.489} \\
crx               &           0.692 &           0.714 &           0.534 &           0.650 &           0.692 &  \textbf{0.725} \\
dermatology       &  \textbf{0.977} &           0.961 &           0.658 &           0.954 &           0.957 &           0.897 \\
german            &           0.516 &           0.513 &           0.504 &  \textbf{0.536} &           0.520 &           0.526 \\
glass             &           0.626 &           0.610 &  \textbf{0.633} &           0.510 &           0.574 &           0.518 \\
haberman          &           0.550 &           0.560 &           0.500 &           0.498 &           0.508 &  \textbf{0.581} \\
heart             &  \textbf{0.809} &           0.798 &           0.551 &  \textbf{0.809} &           0.800 &           0.802 \\
ion        &           0.831 &           0.790 &           0.839 &  \textbf{0.843} &           0.753 &           0.636 \\
mammo      &           0.818 &           0.810 &           0.500 &           0.820 &           0.790 &  \textbf{0.821} \\
page       &           0.630 &           0.612 &           0.597 &           0.520 &           0.590 &  \textbf{0.643} \\
phoneme           &  \textbf{0.847} &           0.843 &           0.401 &           0.832 &           0.821 &           0.840 \\
pima              &           0.687 &           0.703 &           0.500 &           0.668 &  \textbf{0.711} &           0.688 \\
titanic           &           0.682 &           0.682 &           0.664 &           0.682 &  \textbf{0.687} &           0.682 \\
vehicle           &           0.709 &           0.698 &           0.603 &           0.703 &           0.708 &  \textbf{0.797} \\
vowel             &           0.920 &           0.906 &           0.782 &           0.891 &           0.787 &  \textbf{0.921} \\
wdbc              &           0.960 &  \textbf{0.969} &           0.500 &           0.930 &           0.932 &           0.699 \\
wine              &  \textbf{0.977} &  \textbf{0.977} &           0.774 &           0.960 &           0.966 &           0.965 \\
red   &           0.304 &           0.309 &           0.266 &           0.297 &           0.294 &  \textbf{0.324} \\
white &           0.319 &           0.317 &           0.237 &           0.302 &           0.288 &  \textbf{0.340} \\
wisconsin         &           0.958 &  \textbf{0.962} &           0.500 &           0.945 &           0.950 &           0.922 \\
yeast             &           0.547 &           0.555 &           0.182 &           0.470 &           0.552 &  \textbf{0.557} \\
\midrule
mean              &  \textbf{0.693} &           0.689 &           0.514 &           0.677 &           0.667 &           0.669 \\
\bottomrule
\end{tabular}

    \caption{Balanced accuracy of FRNN on the benchmark data sets with the classical relations, \(k = 20\).}
    \label{tab:ap:k20}
\end{table}

\begin{table}[hp]
    \centering
    \begin{tabular}{lrrrrrr}
    \toprule
    data set &     1  &     3  &     5  &     10 &     15 &     20 \\
    \midrule
    aust    &  0.808 &  0.834 &  0.838 &  0.852 &  0.852 &  0.852 \\
    bands         &  0.656 &  0.685 &  0.655 &  0.679 &  0.645 &  0.633 \\
    bupa          &  0.606 &  0.635 &  0.649 &  0.672 &  0.669 &  0.676 \\
    contra &  0.444 &  0.460 &  0.470 &  0.492 &  0.493 &  0.501 \\
    crx           &  0.653 &  0.700 &  0.734 &  0.739 &  0.753 &  0.747 \\
    german        &  0.537 &  0.541 &  0.560 &  0.546 &  0.532 &  0.538 \\
    haberman      &  0.620 &  0.626 &  0.596 &  0.592 &  0.581 &  0.580 \\
    heart         &  0.773 &  0.791 &  0.802 &  0.815 &  0.830 &  0.834 \\
    ion    &  0.851 &  0.875 &  0.857 &  0.834 &  0.827 &  0.820 \\
    mammo  &  0.750 &  0.808 &  0.822 &  0.830 &  0.830 &  0.828 \\
    pima          &  0.632 &  0.681 &  0.674 &  0.686 &  0.699 &  0.717 \\
    sonar         &  0.851 &  0.870 &  0.867 &  0.891 &  0.840 &  0.855 \\
    spectfheart   &  0.643 &  0.597 &  0.586 &  0.627 &  0.593 &  0.619 \\
    titanic       &  0.521 &  0.538 &  0.631 &  0.631 &  0.682 &  0.682 \\
    vehicle       &  0.759 &  0.756 &  0.765 &  0.777 &  0.745 &  0.766 \\
    vowel         &  0.994 &  0.987 &  0.985 &  0.974 &  0.960 &  0.946 \\
    wdbc          &  0.961 &  0.969 &  0.973 &  0.969 &  0.977 &  0.972 \\
    wine          &  0.984 &  0.984 &  0.980 &  0.975 &  0.980 &  0.980 \\
    wisconsin     &  0.944 &  0.960 &  0.965 &  0.971 &  0.971 &  0.972 \\
    \midrule
    mean          &  0.736 &  0.752 &  0.758 &  0.766 &  0.761 &  0.764 \\
    \bottomrule
    \end{tabular}
    \caption{The balanced accuracy of FRNN with NCA for \(k \in \{1,3,5,10,15,20\}.\) }
    \label{tab:ap:ncasens}
\end{table}

\begin{table}
    \centering
    \begin{tabular}{lllllll}
    \toprule
    data set &     1  &     3  &     5  &     10 &     15 &     20 \\
    \midrule
    aust    &  0.816 &  0.848 &  0.848 &  0.854 &  0.859 &  0.863 \\
    bands         &  0.711 &  0.712 &  0.695 &  0.662 &  0.640 &  0.628 \\
    bupa          &  0.613 &  0.647 &  0.656 &  0.672 &  0.670 &  0.664 \\
    contra &  0.414 &  0.436 &  0.454 &  0.462 &  0.468 &  0.468 \\
    crx           &  0.645 &  0.685 &  0.700 &  0.695 &  0.688 &  0.692 \\
    german        &  0.570 &  0.549 &  0.548 &  0.527 &  0.523 &  0.516 \\
    haberman      &  0.536 &  0.558 &  0.530 &  0.529 &  0.550 &  0.550 \\
    heart         &  0.781 &  0.807 &  0.814 &  0.808 &  0.806 &  0.809 \\
    ion    &  0.878 &  0.871 &  0.859 &  0.855 &  0.851 &  0.831 \\
    mammo  &  0.746 &  0.802 &  0.812 &  0.816 &  0.816 &  0.818 \\
    pima          &  0.656 &  0.671 &  0.674 &  0.676 &  0.691 &  0.687 \\
    sonar         &  0.852 &  0.849 &  0.864 &  0.845 &  0.843 &  0.838 \\
    spectfheart   &  0.614 &  0.605 &  0.664 &  0.637 &  0.616 &  0.612 \\
    titanic       &  0.521 &  0.532 &  0.631 &  0.631 &  0.681 &  0.682 \\
    vehicle       &  0.694 &  0.709 &  0.714 &  0.713 &  0.709 &  0.709 \\
    vowel         &  0.994 &  0.987 &  0.985 &  0.959 &  0.939 &  0.920 \\
    wdbc          &  0.951 &  0.964 &  0.966 &  0.962 &  0.964 &  0.960 \\
    wine          &  0.958 &  0.967 &  0.967 &  0.967 &  0.977 &  0.977 \\
    wisconsin     &  0.956 &  0.965 &  0.967 &  0.966 &  0.962 &  0.958 \\
    \midrule
    mean          &  0.732 &  0.746 &  0.755 &  0.749 &  0.750 &  0.746 \\
    \bottomrule
    \end{tabular}
    \caption{The balanced accuracy of FRNN with the Manhattan distance for \(k \in \{1,3,5,10,15,20\}.\) }
    \label{tab:ap:mansens}
\end{table}

\clearpage
\newpage


\bibliographystyle{plain}
\bibliography{sources}

\end{document}